\documentclass{ar-1col-S2O}
\usepackage[numbers]{natbib}
\usepackage{url}
\usepackage{array}
\usepackage{booktabs}
\usepackage{graphicx}
\usepackage[hidelinks]{hyperref}
\usepackage{lipsum}  
\usepackage{comment}
\usepackage{multirow, makecell}
\usepackage{subcaption}
\usepackage[titletoc]{appendix}

\jname{Xxxx. Xxx. Xxx. Xxx.}
\jvol{AA}
\jyear{YYYY}
\doi{10.1146/((please add article doi))}

\newcolumntype{P}[1]{>{\centering\arraybackslash}p{#1}}
\newcolumntype{R}[1]{>{\raggedleft\arraybackslash}p{#1}}
\UseRawInputEncoding

\begin{document}

\markboth{Tang C., Abbatematteo B., Hu J., Chandra R., Mart{\'i}n-Mart{\'i}n R., Stone P.}{Real-World Successes of DRL in Robotics}
\title{Deep Reinforcement Learning for Robotics: \\A Survey of Real-World Successes}

\author{Chen Tang{$^{1,*}$}, Ben Abbatematteo{$^{1,*}$}, \\Jiaheng Hu{$^{1,*}$}, Rohan Chandra{$^{2}$}, \\Roberto Mart{\'i}n-Mart{\'i}n{$^1$}, Peter Stone{$^{1,3}$}
\affil{$^1$Department of Computer Science, The University of Texas at Austin, Austin, Texas 78712, United States; email: chen.tang@utexas.edu, abba@cs.utexas.edu, jiahengh@utexas.edu, robertomm@cs.utexas.edu, pstone@utexas.edu}
\affil{$^2$Department of Computer Science, The University of Virginia, Charlottesville, Virginia 22904, United States; email:  rohanchandra@virginia.edu}
\affil{$^3$Sony AI}
\affil{$^*$Equal Contribution}
}

\newcommand{\commentp}[1]{{\color{green}{\bf [Peter: #1]}}}
\newcommand{\banote}[1]{{\color{cyan}{\bf [Ben: #1]}}}
\newcommand{\commentr}[1]{{\color{orange}{\bf [Rohan: #1]}}}
\newcommand{\jiaheng}[1]{{\color{blue}{\bf [Jiaheng: #1]}}}
\newcommand{\chen}[1]{{\color{red}{\bf [Chen: #1]}}}
\newcommand{\roberto}[1]{{\color{magenta}{\bf [Roberto: #1]}}}


\definecolor{L0}{rgb}{0.99, 0.99, 1.0}
\definecolor{L1}{rgb}{0.8, 0.95, 0.95}
\definecolor{L2}{rgb}{0.529, 0.808, 0.962}
\definecolor{L3}{rgb}{0.000, 0.700, 0.745}
\definecolor{L4}{rgb}{0.118, 0.565, 1.000}
\newcommand{\Level}[2]{\colorbox{L#1}{\textcolor{black}{\citealp{#2}}}}

\begin{abstract}
Reinforcement learning (RL), particularly its combination with deep neural networks referred to as deep RL (DRL), has shown tremendous promise across a wide range of applications, suggesting its potential for enabling the development of sophisticated robotic behaviors. Robotics problems, however, pose fundamental difficulties for the application of RL, stemming from the complexity and cost of interacting with the physical world. 
This article provides a modern survey of DRL for robotics, with a particular focus on evaluating the real-world successes achieved with DRL in realizing several key robotic competencies. Our analysis aims to identify the key factors underlying those exciting successes, reveal underexplored areas, and provide an overall characterization of the status of DRL in robotics. We highlight several important avenues for future work, emphasizing the need for stable and sample-efficient real-world RL paradigms, holistic approaches for discovering and integrating various competencies to tackle complex long-horizon, open-world tasks, and principled development and evaluation procedures. This survey is designed to offer insights for both RL practitioners and roboticists toward harnessing RL's power to create generally capable real-world robotic systems.
\end{abstract}

\maketitle



\section{Introduction}
Reinforcement learning (RL)~\cite{sutton2018reinforcement} refers to a class of decision-making problems in which an agent must learn through trial-and-error to act in such a way that maximizes its accumulated \emph{return}, as encoded by a scalar reward function that maps the agent's states and actions to immediate rewards. 
RL algorithms, particularly their combination with deep neural networks referred to as deep RL (DRL)~\cite{franccois2018introduction}, have shown 
 remarkable capabilities in solving complex decision-making problems even with high-dimensional observations in domains such as board games~\cite{schrittwieser2020mastering}, video games~\cite{wurman2022outracing}, healthcare~\cite{yu2021reinforcement}, and recommendation systems~\cite{afsar2022reinforcement}. 

These successes underscore the potential of DRL for controlling robotic systems with high-dimensional state or observation space and highly nonlinear dynamics to perform challenging tasks that conventional decision-making, planning, and control approaches (e.g., classical control, optimal control, sampling-based planning) cannot handle effectively. Yet, the most notable milestones of DRL so far have been achieved in simulation or game environments, where RL agents can learn from extensive experience. In contrast, robots need to complete tasks in the \emph{physical world}, which presents additional challenges. It is often inefficient and/or unsafe for the RL agents to collect trial-and-error samples directly in the physical world, and it is usually impossible to create an exact replica of the complex real world in simulation. These challenges notwithstanding, recent advances have enabled DRL to succeed at some real-world robotic tasks. For instance, DRL has enabled champion-level drone racing~\cite{kaufmann2023champion} and versatile quadruped locomotion control integrated into production-level quadruped systems (e.g., ANYbotics\footnote{\url{https://www.anybotics.com/news/superior-robot-mobility-where-ai-meets-the-real-world/}}, Swiss-Mile\footnote{\url{https://www.swiss-mile.com/}}, and Boston Dynamics\footnote{\url{https://bostondynamics.com/blog/starting-on-the-right-foot-with-reinforcement-learning/}}). However, \emph{the maturity of state-of-the-art DRL solutions varies significantly across different robotic applications}. In some domains, such as urban autonomous driving, DRL-based solutions remain limited to simulation or strictly confined field tests~\cite{kiran2021deep}.  

This survey aims to comprehensively evaluate the current progress of DRL in real-world robotic applications, identifying key factors behind the most exciting successes and open challenges that remain in less mature areas. Specifically, we assess the maturity of DRL for a variety of problem domains and contrast the DRL literature across domains to pinpoint broadly applicable techniques, under-explored areas, and common open challenges that need to be addressed to advance DRL's applications in robotics. We aim for this survey to provide researchers and practitioners with a thorough understanding of the status of DRL in robotics, offering valuable insights to guide future research and facilitate broadly deployable DRL solutions for real-world robotic tasks.

\section{Why Another Survey on RL for Robotics?}
Although some previous articles have surveyed RL for robotics, we make three contributions that provide unique perspectives on the literature and fill gaps in knowledge. First, we focus on work that has demonstrated at least \emph{some degree of real-world success}, aiming to assess the current state and open challenges of DRL for real-world robotic applications. Most existing surveys on RL for robotics do not explicitly address this topic, e.g., Dulac-Arnold et al.~\cite{dulac-arnold_challenges_2019} discuss the general challenges of real-world RL not specific to robotics, and Ibarz et al.~\cite{ibarz_how_2021} list open challenges of DRL unique to real-world robotics settings but based on case studies drawn only from their own research. In contrast, our discussion is grounded in a comprehensive assessment of the real-world successes achieved by DRL in robotics, with one aspect of our evaluation being the level of real-world deployment (see Sec.~\ref{sec:taxonomy-levels}). 

Second, we present a \emph{novel} and \emph{comprehensive} taxonomy that categorizes DRL solutions along multiple axes: robot competencies learned with DRL, problem formulation, solution approach, and level of real-world success. Prior surveys on RL for robotics and broader robot learning have often focused on specific tasks~\cite{kroemer_review_2019,xiao2022motion} or on particular techniques~\cite{deisenroth_survey_2011,brunke_safe_2022}. By contrast, our taxonomy allows us to survey the complete landscape of DRL solutions that are effective in robotics application domains, in addition to reviewing the literature of each application domain separately. Within this framework, we compare and contrast solutions and identify \emph{common patterns, broadly applicable approaches, under-explored areas, and open challenges} for realizing successful robotic systems. 

Third, while some past surveys have shared our motivation to provide a broad analysis of the field, the fast and impressive pace of DRL progress has created the need for a renewed analysis of the field, its successes, and limitations. The seminal survey by Kober et al.~\cite{kober_reinforcement_2013} was written before the deep learning era, and the general deep learning for robotics survey by Sunderhauf et al.~\cite{sunderhauf2018limits} was written when DRL accomplishments were primarily in simulation. We provide a refreshed overview of the field by focusing on DRL, which is behind the most notable real-world successes of RL in robotics, paying particular attention to papers published in the last five years, during which most of the successes occurred. 
\section{Taxonomy}
\label{ss:taxonomy}
\begin{figure}[t]
    \centering
    \includegraphics[width=0.99\textwidth]{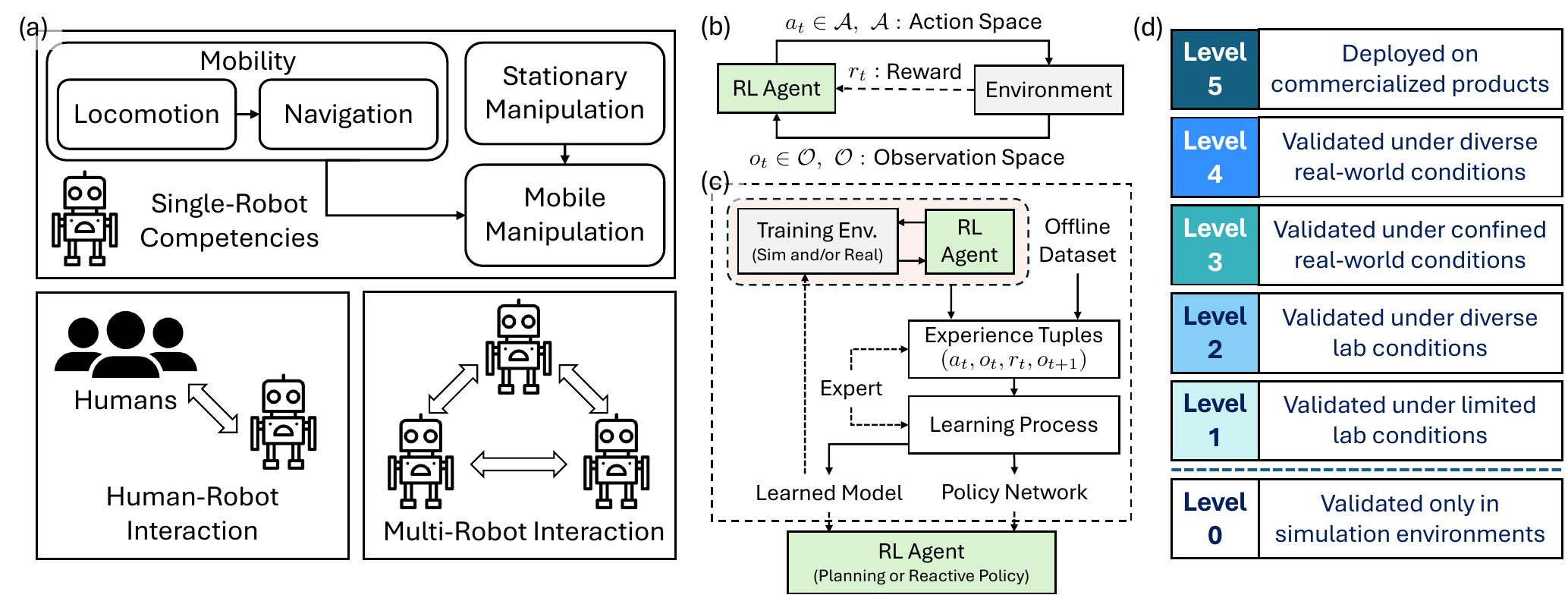}
    \caption{The four aspects of our taxonomy: (a) Robot competencies learned with DRL; (b) Problem formulation; (c) Solution approach; and (d) Levels of real-world success.}
    \label{fig:taxonomy}
\end{figure}

This section presents the novel taxonomy we introduce to categorize the literature on DRL. The unique focus of our survey on the real-world successes of DRL in robotics necessitates a new taxonomy to categorize and analyze the literature, which should enable us to assess the maturity of DRL solutions across various robotic applications and derive valuable lessons from both successes and failures. Specifically, we should identify the specific robotic problem addressed in each paper, understand how it has been abstracted as an RL problem, and summarize the DRL techniques applied to solve it. More importantly, we should evaluate the maturity of these DRL solutions, as demonstrated in their experiments. Consequently, we introduce a taxonomy spanning four axes: \textbf{robot competencies learned with DRL}, \textbf{problem formulation}, \textbf{solution approach}, and \textbf{the level of real-world success}.

\subsection{Robot Competencies Learned with DRL}\label{sec:robot-application}
Our primary axis focuses on the target robotic task studied in each paper. A robotic task, especially in open real-world scenarios, may require multiple competencies. One may apply DRL to synthesize an end-to-end system to realize all the competencies or learn sub-modules to enable a subset of them. Since our focus is DRL, we classify papers based on \emph{the specific robot competencies learned and realized with DRL}. We first classify the competencies into \emph{single-robot}\textemdash competencies required for a robot to complete tasks on its own\textemdash and \emph{multi-agent}\textemdash competencies required to interact with other agents sharing the workspace with the robot and affecting its task completion. 

When a single robot completes a task in a workspace, any competencies it requires can be considered as enabling specific ways to \emph{interact with and affect the physical world}, which are further divided into \texttt{mobility}\textemdash moving in the environment\textemdash and \texttt{manipulation}\textemdash moving or rearranging (e.g., grasping, rotating) objects in the environment~\cite{mason2001mechanics,siciliano2008springer,mason2018toward}. In the robotics literature, mobility\footnote{In the robotics literature, both \texttt{locomotion} and \texttt{navigation} have been used to refer to the ability to move in an environment. To avoid confusion, \texttt{mobility} is used in this survey to refer to the overarching category where DRL enables robot movement.} is typically split into two problems: \texttt{locomotion} and \texttt{navigation}~\cite{siciliano2008springer,rudin2022advanced}. Locomotion focuses on motor skills that enable robots of various morphologies (e.g., quadrupeds, humanoids, wheeled robots, drones) to traverse different environments, while navigation focuses on strategies that direct a robot to its destination efficiently without collision. Typical navigation policies generate \emph{high-level} motion commands, such as desired states at the center of mass (CoM), while assuming effective locomotion control to execute them~\cite{siciliano2008springer}. Some works jointly address the locomotion and navigation problems, which is particularly useful for tasks in which the navigation strategies are heavily affected by the robot's capability to traverse the environment, as determined by the robot dynamics and locomotion control (e.g., navigating through challenging terrains~\cite{rudin2022advanced} or racing~\cite{song2023reaching}). We review these papers alongside other navigation papers since their ultimate goal is navigation. 

In the robotics literature, manipulation is often studied in table-top settings, e.g., robotic arms or hands mounted on a stationary base with fixed sensors observing the scene. Some other real-world tasks further require robots to interact with the environment while moving their base (e.g., household and warehouse robots), which necessitates a synergistic integration of manipulation and mobility capabilities. We review the former case under the \texttt{stationary manipulation} category and the latter under \texttt{mobile manipulation}. 

When the task completion is affected by the other agents in the workspace, the robot needs to be further equipped with \emph{abilities to interact with other agents}, which we place under the heading of \emph{multi-agent} competencies. Note that some single-robot competencies may still be required while the robot interacts with others, such as crowd navigation or collaborative manipulation. In this category, we focus on papers where DRL occurs at the agent-interaction level, i.e., learning interaction strategies given certain single-robot competencies or learning policies that jointly optimize interaction and single-robot competencies. We further split these works into two subcategories based on the types of agents the robot interacts with: 1) \emph{Human-robot interaction} concerns a robot's ability to operate alongside humans. The presence of humans introduces additional challenges due to their sophisticated behavior and the stringent safety requirements for robots operating around humans.~2) \emph{Multi-robot interaction} refers to a robot's ability to interact with a group of robots. A class of RL algorithms, multi-agent RL (MARL), is typically applied to solve this problem. In MARL, each robot is a learning agent evolving its policy based on its interactions with the environment and other robots, which complicates the learning mechanism. Depending on whether the robots' objectives align, their interactions could be cooperative, adversarial, or general-sum. In addition, practical scenarios often require decentralized decision-making under partial observability and limited communication bandwidth.  

\subsection{Problem Formulation} The second axis of our taxonomy is the formulation of the RL problem, which specifies the optimal control policy for the targeted robot competency. RL problems are typically modeled as Partially Observable Markov Decision Processes (POMDPs) for single-agent RL and Decentralized POMDPs (Dec-POMDP) for multi-agent RL. Specifically, we categorize the papers based on the following elements of the problem formulation: 1) \emph{Action space}: whether the actions are \emph{low-level} (i.e., joint or motor commands), \emph{mid-level} (i.e., task-space commands), or \emph{high-level} (i.e., temporally extended task-space commands or subroutines); 2) \emph{Observation space}: whether the observations are \emph{high-dimensional} sensor inputs (e.g., images and/or LiDAR scans) or estimated \emph{low-dimensional} state vectors; 3) \emph{Reward function}: whether the reward signals are \emph{sparse} or \emph{dense}. Due to space limitations, we provide detailed definitions of these terms in the supplementary materials.

\subsection{Solution Approach} Another axis closely related to the previous one is the solution approach used to solve the RL problem, which is composed of the RL algorithm and associated techniques that enable a practical solution for the target robotic problem. Specifically, we classify the solution approach from the following perspectives: 1) \emph{Simulator usage}: whether and how simulators are used, categorized into \emph{zero-shot}, \emph{few-shot sim-to-real transfer}, or directly learning offline or in the real world \emph{without simulators}; 2) \emph{Model learning}: whether (a part of) the transition dynamics model is learned from robot data; 3) \emph{Expert usage}: whether expert (e.g., human or oracle policy) data are used to facilitate learning; 4) \emph{Policy optimization}: the policy optimization algorithm adopted, including \emph{planning} or \emph{offline}, \emph{off-policy}, or \emph{on-policy RL}; 5) \emph{Policy/Model Representation}: Classes of neural network architectures used to represent the policy or dynamics model, including \emph{MLP}, \emph{CNN}, \emph{RNN}, and \emph{Transformer}. Please refer to the supplementary materials for detailed term definitions.

\subsection{Level of Real-World Success}\label{sec:taxonomy-levels} To evaluate the practicality of DRL in real-world robotic tasks, we categorize the papers based on the maturity of their DRL methods. By comparing the effectiveness of DRL across different robotic tasks, we aim to identify domains where the gaps between research prototypes and real-world deployment are more or less significant. This requires a metric to quantify real-world success across tasks, which, to our knowledge, has not been attempted in the DRL for robotics literature. Inspired by the levels of autonomous driving~\cite{sae} and Technology readiness level (TRL) for machine learning~\cite{lavin2022technology}, we introduce the concept of \emph{levels of real-world success}. We classify the papers into six levels based on the scenarios where the proposed methods have been validated: 1) \emph{Level 0}: validated only in simulation; 2) \emph{Level 1}: validated in limited lab conditions; 3) \emph{Level 2}: validated in diverse lab conditions; 4) \emph{Level 3}: validated under confined real-world operational conditions; 5) \emph{Level 4}: validated under diverse, representative real-world operational conditions; and 6) \emph{Level 5}: deployed on commercialized products. We consider Levels 1-5 as achieving at least some degree of real-world success.
The only information we can use to assess the level of real-world success is the experiments reported by the authors. However, many papers only described a single real-world trial. While we strive to provide accurate estimates, this assessment can be subjective due to limited information. Additionally, we use the level of real-world success to quantify the maturity of a solution for its target problem, irrespective of its complexity. 



\section{Competency-Specific Review}\label{sec:review}
This section provides a detailed review of the DRL literature, with each subsection focusing on a specific robot competency. In each subsection, we further organize the review based on subcategories specific to each type of competency. After discussing the papers, we conclude each subsection by summarizing the trends and open challenges for learning the competency in question. To aid understanding, each subsection includes a table to overview the reviewed papers. Since our main objective is to assess the maturity of DRL solutions, we note the level of real-world success achieved by each paper in the table. For a comprehensive categorization of the papers, please refer to Tables~\ref{tab:taxonomy-mdp}--\ref{tab:taxonomy-solution4} in the supplementary materials.

\subsection{Locomotion}\label{sec:locomotion}
\begin{figure*}[t!]
    \centering
    \begin{subfigure}[c]{0.4\textwidth}
        \centering
        \includegraphics[width=0.9\textwidth]{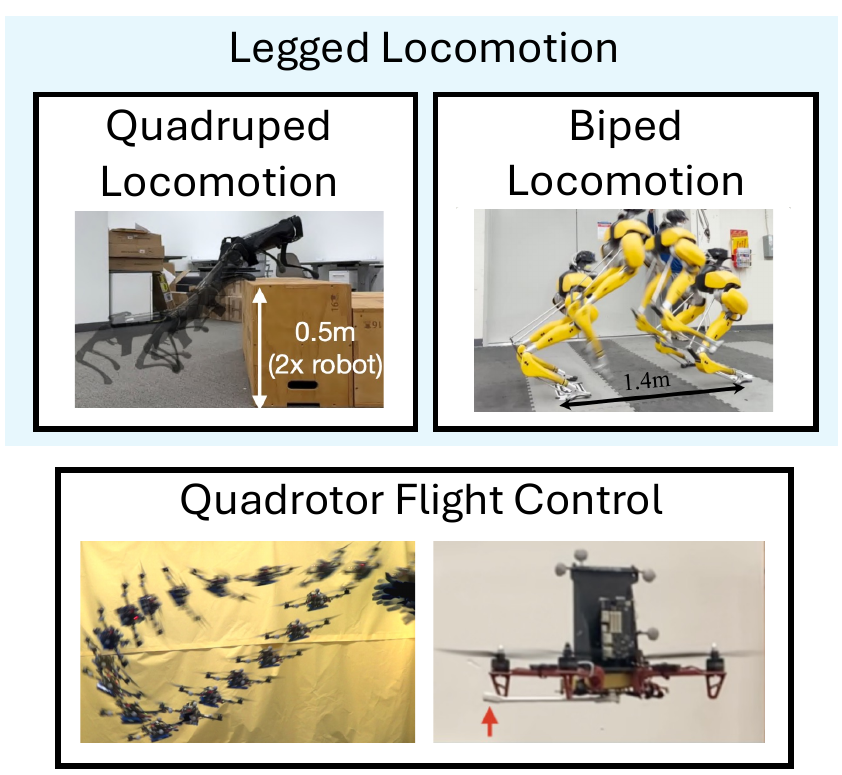}
    \end{subfigure}
    \begin{subfigure}[c]{0.5\textwidth}
        \centering 
                \begin{tabular}{P{1.6cm}|P{4.2cm}}
        \toprule 
        Quadruped & \Level{1}{tan2018sim}, \Level{1}{hwangbo2019learning}, \Level{2}{feng2023genloco}, \Level{2}{lee2019robust}, \Level{3}{yang2020multi}, \Level{4}{kumar2021rma}, \Level{4}{lee2020learning}, \Level{4}{miki2022learning}, \Level{4}{gangapurwala2022rloc}, \Level{3}{choi2023learning}, \Level{4}{nahrendra2023dreamwaq}, \Level{1}{escontrela2022adversarial}, \Level{1}{ma2023learning}, \Level{3}{fu2022minimizing}, \Level{4}{loquercio2023learning}, \Level{4}{agarwal2023legged}, \Level{4}{yang2023neural}, \Level{4}{jenelten2024dtc}, \Level{3}{yang2023cajun}, \Level{3}{smith2022legged}, \Level{3}{cheng2023parkour}, \Level{3}{zhuang2023robot}, \Level{3}{vollenweider2023advanced}, \Level{3}{margolis2023walk}, \Level{3}{smith2023demonstrating}, \Level{1}{wu2023daydreamer}\\ \hline 
        
        Biped & \Level{2}{kumar2022adapting}, \Level{1}{siekmann2020learning}, \Level{1}{hanna2021grounded}, \Level{1}{siekmann2021sim}, \Level{3}{li2021reinforcement}, \Level{3}{siekmann2021blind}, \Level{3}{castillo2022reinforcement}, \Level{2}{duan2023learning}, \Level{3}{radosavovic2023real}, \Level{3}{li_reinforcement_2024} \\ \hline
        
        Flight & \Level{1}{hwangbo2017control}, \Level{2}{molchanov2019sim}, \Level{3}{kaufmann2022benchmark}, \Level{2}{zhang2023hover}, \Level{2}{eschmann2024learning}\\ \bottomrule
        \end{tabular}
        \vspace{8pt}
    \end{subfigure}
    \caption{\textbf{Left:} An overview of the three locomotion problems reviewed in Sec.~\ref{sec:locomotion}, including quadruped~\cite{cheng2023parkour} and biped~\cite{li_reinforcement_2024} locomotion, and quadrotor flight control~\cite{hwangbo2017control,zhang2023hover}; \textbf{Right:} Locomotion papers reviewed in Sec.~\ref{sec:locomotion}. The color map indicates the levels of real-world success: \colorbox{L1}{\emph{\textcolor{black}{Limited Lab}}}, \colorbox{L2}{\emph{\textcolor{black}{Diverse Lab}}}, \colorbox{L3}{\emph{\textcolor{black}{Limited Real}}}, and \colorbox{L4}{\emph{\textcolor{black}{Diverse Real}}}.}\label{fig:locomotion} 
    \vspace{-4pt}
\end{figure*}

Locomotion research aims to develop motor skills for robots to traverse various real-world environments. Prior to the deep learning era, several pioneering works have explored RL for locomotion control and delivered promising hardware demos, e.g., quadruped walking~\cite{kohl2004policy} and helicopter control~\cite{bagnell2001autonomous,abbeel2006application}. This subsection reviews DRL solutions for locomotion separately from navigation, where the controllers follow high-level navigation commands. Since locomotion mainly concerns motor skills, the problem complexity is primarily influenced by the system dynamics~\cite{kumar2022adapting}. We organize this subsection accordingly and review three representative locomotion problems: \textbf{quadruped and biped locomotion}, and \textbf{quadrotor flight control}. See Figure~\ref{fig:locomotion} for an overview of the papers reviewed. 

\subsubsection{Quadruped Locomotion}\label{sec:quadruped}
Quadruped locomotion is one of the robotic domains where DRL has provided mature real-world solutions. Multiple robotics companies, such as ANYbotics, Swiss-Mile, and Boston Dynamics, have reported that DRL was integrated into their quadruped control for applications including industrial inspection, last-mile delivery, and rescue operations. In the literature, DRL methods were first validated for \emph{blind quadruped walking}, i.e., relying solely on proprioceptive sensors on flat indoor surfaces~\cite{tan2018sim,hwangbo2019learning}. These policies were typically trained in simulation and deployed zero-shot in the real world. The main challenge lies in the sim-to-real gap in quadrupeds' intrinsic dynamics. Several strategies have been explored to bridge the reality gap: 1) learning actuator models, either analytical~\cite{tan2018sim} or neural network-based~\cite{hwangbo2019learning}, from robot data to improve simulation fidelity; 2) randomizing dynamics parameters~\cite{tan2018sim,hwangbo2019learning} and, even further, randomizing morphology~\cite{feng2023genloco}, which enables generalization to unseen quadrupeds; and 3) adopting a hierarchical structure with a low-level, model-based controller to handle dynamics discrepancy and external disturbances while facilitating efficient learning. The interface between the DRL policy and the model-based controller could be defined at various levels, such as joint positions~\cite{lee2019robust,yang2020multi,kumar2021rma}, leg poses~\cite{tan2018sim}, gait parameters~\cite{lee2020learning,miki2022learning}, or temporally-extended macro actions~\cite{gangapurwala2022rloc}. 
As robots venture beyond controlled lab environments, they encounter more challenging terrains such as discontinuous, deformable, or slippery surfaces. Four main techniques have been used to address the additional challenges. First, the terrain and contact information are not directly observable. Privileged learning has been commonly adopted as a solution~\cite{lee2020learning,kumar2021rma}, where a policy with privileged terrain information is trained first and then distilled into a student policy operating on realistic sensor inputs. Alternatively, end-to-end training can be achieved with the help of state estimation~\cite{choi2023learning,nahrendra2023dreamwaq} and asymmetric actor-critic~\cite{pinto_asymmetric_2018,nahrendra2023dreamwaq}. In both cases, an extended history of observations is often set as input. 

Second, policies should be exposed to diverse conditions during training for generalization in the wild. A learning curriculum that progressively increases task difficulty is often adopted to facilitate training~\cite{lee2020learning,kumar2021rma,gangapurwala2022rloc,nahrendra2023dreamwaq,choi2023learning}. Advanced terrain models can also improve performance on terrains with complex contact dynamics, e.g., deformable surfaces~\cite{choi2023learning}.

Third, exteroceptive sensors are crucial for traversing risky terrains, as they allow the quadruped to adapt to terrains without stepping on them. For example, they have fostered more efficient and robust stair traversal~\cite{miki2022learning,loquercio2023learning}. Exteroceptive observations are typically in the form of terrain height maps~\cite{miki2022learning,gangapurwala2022rloc}, depth images~\cite{agarwal2023legged,yang2023neural}, and RGB images~\cite{loquercio2023learning}. Privileged learning is widely used to facilitate policy learning from these high-dimensional observations~\cite{agarwal2023legged,miki2022learning,yang2023neural}. To reduce the sim-to-real gap in sensor inputs, techniques such as injecting simulated sensor noise~\cite{miki2022learning}, post-processing depth images~\cite{zhuang2023robot}, learning vision encoders from real-world samples~\cite{loquercio2023learning} are shown effective. Additionally, policies benefit from improved representation via self-supervised learning~\cite{gangapurwala2022rloc,miki2022learning}, cross-modal embedding matching~\cite{agarwal2023legged,loquercio2023learning}, or using models with higher capacity, such as transformers~\cite{yang2021learning,yang2023neural}. 

Fourth, traversing certain complex terrains demands advanced locomotion skills beyond regular walking gaits. For example, end-to-end DRL policies typically struggle with terrains that have sparse contact regions. Jenelten et al.~\cite{jenelten2024dtc} showed that training an RL policy to track reference footholds provided by trajectory optimization results in more accurate and robust foot placement on sparse terrains. Jumping further extends the robots' ability to cross gaps beyond their body length. For example, Yang et al.~\cite{yang2023cajun} trained a DRL policy to generate trajectories with a model-based tracking controller handling the complex jumping dynamics. Fall recovery is another essential skill, especially for automatic reset in real-world RL~\cite{smith2022legged,smith2023demonstrating}. Several works have trained DRL policies for fall recovery~\cite{hwangbo2019learning,lee2019robust,smith2022legged,yang2020multi,ma2023learning}. However, both jumping and fall recovery have only been validated on flat surfaces so far. 

To effectively leverage agile locomotion skills for complex downstream tasks like parkour~\cite{cheng2023parkour,zhuang2023robot}, it is crucial to develop \emph{multi-skill} policies. Learning multiple skills jointly has also been shown effective in fostering policy robustness~\cite{li_reinforcement_2024}. One approach is to create a set of RL policies~\cite{vollenweider2023advanced,yang2020multi,zhuang2023robot}, each tailored to a specific skill, and then train a high-level policy to select the optimal skill~\cite{yang2020multi}. Alternatively, a single policy can be distilled from specialized skill policies through BC~\cite{zhuang2023robot}. To avoid the cumbersome procedure of training multiple specialized policies, several works explored constructing a unified policy directly. For instance, MoB~\cite{margolis2023walk} encoded various locomotion strategies into a single policy conditioned on gait parameters. Cheng et al.~\cite{cheng2023parkour} used a unified reward consisting of waypoint and velocity tracking terms to learn diverse parkour skills. Fu et al.~\cite{fu2022minimizing} showed that energy minimization led to smooth gait transitions. Motion imitation reward is another widely used and unified approach for learning naturalistic and diverse locomotion skills~\cite{vollenweider2023advanced,escontrela2022adversarial}. 

{\bf Remark on RL algorithms.}We conclude the review on quadruped locomotion with a remark on the RL algorithms used in the literature. The most mature DRL solutions for quadruped locomotion followed the zero-shot sim-to-real transfer scheme, predominantly using on-policy model-free RL, e.g., PPO~\citep{schulman2017proximal}, due to its robustness to hyperparameters. Gangapurwala et al.~\cite{gangapurwala2022rloc} noted that on-policy RL could be less favorable when the action space is temporally extended or deterministic control actions are preferred. Meanwhile, researchers have explored few-shot adaptation and real-world RL, either model-free~\cite{smith2022legged,smith2023demonstrating} or model-based~\cite{wu2023daydreamer}, to update policies using real-world rollouts to further generalize policies to novel situations without accurate simulation. However, most works along this line have only been validated in limited lab settings. The state-of-the-art performance for real-world fine-tuning~\cite{smith2022legged} and learning from scratch~\cite{smith2023demonstrating} were achieved by using off-policy RL to learn walking and fall recovery. However, the tested conditions remain limited compared to mature zero-shot solutions. 

\subsubsection{Biped Locomotion} Compared to the quadruped case, the DRL literature on bipedal locomotion is sparser, and the real-world capabilities demonstrated are more limited. We confine the discussion to 3D bipedal robots, which can move freely in all spatial dimensions, unlike 2D bipeds that are attached to booms and confined to 2D planar motion~\cite{grizzle2009mabel}, for their greater practical utility. The literature begins with walking on flat indoor surfaces~\cite{siekmann2020learning,siekmann2021sim} and extends to walking on various indoor~\cite{hanna2021grounded,li2021reinforcement,kumar2022adapting} and outdoor terrains~\cite{castillo2022reinforcement,radosavovic2023real}, and under external forces~\cite{kumar2022adapting,li2021reinforcement}. Other demonstrated skills include stair traversal~\cite{siekmann2021blind}, hopping~\cite{siekmann2021sim}, running~\cite{siekmann2021sim,li_reinforcement_2024}, jumping~\cite{li_reinforcement_2024}, and traversing obstacles and gaps~\cite{duan2023learning}. More advanced skills have been showcased by industrial companies\footnote{For example, Unitree (\url{https://t.ly/s1FwW}) and Boston Dynamics (\url{https://t.ly/NaSaO})}, but no technical reports are publicly available to reveal if RL was used in their demos. Notably, some of these works deployed their locomotion policies on humanoid robots~\cite{hanna2021grounded,castillo2022reinforcement,radosavovic2023real} while others on bipedal robots without upper bodies~\cite{siekmann2020learning,li2021reinforcement,siekmann2021blind,kumar2022adapting,duan2023learning,li_reinforcement_2024}.

The DRL techniques for bipedal locomotion largely overlap with those for quadrupeds but show three distinct trends due to the complex and under-actuated dynamics of bipeds. First, learning basic standing and walking skills is already challenging due to bipeds' non-statically stable dynamics~\cite{siekmann2020learning}. Thus, model-based approaches are frequently used to facilitate RL, either by generating reference gaits to guide RL~\cite{siekmann2020learning,li2021reinforcement,li_reinforcement_2024} or handling low-level control for high-level RL policies~\cite{castillo2022reinforcement}. Alternatively, Siekmann et al.~\cite{siekmann2021sim} offered an end-to-end solution with a reference-free periodic reward design based on periodic composition. Second, the role of state and action \emph{memories} was particularly noted~\cite{siekmann2020learning}, especially a combination of both long- and short-term memories~\cite{li_reinforcement_2024}. Thus, most works adopted sequence models in their policy architecture~\cite{li_reinforcement_2024,duan2023learning,siekmann2020learning,siekmann2021sim,siekmann2021blind,radosavovic2023real}. Third, almost all these policies were zero-shot transferred from simulation. One exception is GAT~\cite{hanna2021grounded}, which collected real-world samples to refine a simulator iteratively, enabling an NAO to walk on uneven carpets. The limited real-world learning examples are likely due to bipeds' limited recovery capabilities, which hinder their resilience in trials, particularly their ability to auto-reset. 

\subsubsection{Quadrotor Flight Control} Flight control for unmanned aerial vehicles (UAVs), in particular quadrotors, is another problem where DRL has shown compelling performance. Hwangbo et al.~\cite{hwangbo2017control} developed the first DRL quadrotor control policy that was successfully validated on hardware for waypoint tracking and recovery from harsh initialization. Later studies showed that carefully designed simulated dynamics, domain randomization~\cite{molchanov2019sim}, and carefully designed action space, specifically collective thrust and body rates~\cite{kaufmann2022benchmark}, can facilitate policy robustness. Zhang et al.~\cite{zhang2023hover} applied RMA to train a robust near-hover position controller adaptable to unseen disturbances. Eschmann et al.~\cite{eschmann2024learning} introduced the first off-policy RL paradigm for quadrotor control, capable of training a deployable control policy within 18 seconds for waypoint tracking.
In summary, DRL has demonstrated better robustness than classical feedback controllers (e.g., PID) in hovering control~\cite{molchanov2019sim,zhang2023hover}. However, DRL policies tend to have larger tracking errors than carefully designed optimization-based controllers for waypoint tracking~\cite{hwangbo2017control,kaufmann2022benchmark}. Yet the fundamental advantage of RL over optimal control is it enables joint optimization for planning and control~\cite{song2023reaching}, making it an ideal candidate for agile navigation such as racing (see Sec.~\ref{sec:navigation}).  

\subsubsection{Trends and Open Challenges in Locomotion} 
In summary, DRL has shown effectiveness in synthesizing robust and adaptive locomotion controllers for challenging conditions. DRL Techniques used for quadruped, biped, and flight control heavily overlap. For instance, RMA~\cite{kumar2021rma}, initially proposed for quadruped locomotion, has been adapted for both biped~\cite{kumar2022adapting} and quadrotor flight control~\cite{zhang2023hover}. However, the maturity of DRL solutions varies across domains. Quadrupeds can traverse various indoor and outdoor terrains via DRL, while real-world bipedal locomotion skills achieved by DRL are more limited. For quadrotors, most tests remain confined to controlled, obstacle-free indoor environments. Hardware accessibility is a contributing factor. The introduction of low-cost quadrupeds has spurred quadruped research and led to open-sourced and unified software packages.
Conversely, the high cost of bipedal hardware limits extensive real-world testing, though recent advances in humanoid hardware are expected to boost biped research. 
More importantly, the quadruped dynamics are inherently more stable, whereas bipeds and quadrotors are more prone to catastrophic failures under control errors, imposing higher requirements on both robustness and precision of control~\cite{li_reinforcement_2024}. High-speed quadrotor control in outdoor scenarios with complex obstacles further requires the policy to ensure the \emph{long-horizon} feasibility of the closed-loop trajectories~\cite{loquercio2021learning}. End-to-end RL integrating long-horizon planning and short-horizon control shows promise as a solution~\cite{kaufmann2023champion}. In addition to ensuring long-horizon feasibility, integrating locomotion with downstream tasks (e.g., loco-manipulation) is an exciting direction in general, but how to discover skills necessary for downstream tasks remains an open question.

\begin{summary} [Key Takeaways]
    \begin{itemize}
        \item DRL has enabled mature quadruped locomotion control; yet, the maturity of DRL-based solutions for other locomotion problems is lower. 
        \item Hardware accessibility is an important contributing factor. Low-cost and standard hardware platforms would facilitate DRL development.
        \item The inherently complex dynamics of certain locomotion problems present fundamental challenges to the reliable deployment of DRL locomotion controllers.
        \item Even in the mature quadruped locomotion domain, open questions remain, such as 1) effectively integrating locomotion with downstream tasks via RL, and 2) enabling efficient and safe real-world learning.  
    \end{itemize}
    \vspace{-8pt}
\end{summary}

\subsection{Navigation}
\label{sec:navigation}
Navigation focuses on the decision-making challenge in mobility: transporting an agent to a goal location while avoiding collisions, typically assuming effective locomotion. 
As a fundamental mobility capability, navigation has an extensive history in robotics research~\cite{siciliano2008springer}. ``Classical" navigation approaches employ mapping, localization, and planning modules to determine and execute a path to a goal. Planning is typically decomposed into global planning, which produces a coarse path, and local planning, which tracks the global plan and handles collision avoidance. In this section, we delineate navigation works by embodiment: \textbf{wheeled, legged, and aerial navigation} and identify capabilities enabled by RL in each setting. Social navigation, where the robot navigates in the presence of humans, is deferred to Sec.~\ref{sec:hri}. Multi-robot navigation is similarly deferred to Sec.~\ref{subsec: multi-robot-interaction}.

\subsubsection{Wheeled Navigation}
Navigation for wheeled robots, in particular, has a long history in robotics~\cite{siciliano2008springer}. We discuss several common wheeled navigation settings, including geometric navigation, visual navigation, and offroad navigation. 

\textbf{Geometric Navigation.}
Early attempts aimed to verify RL's capability in solving navigation problems typically solved with modular classical approaches~\cite{tai_virtual--real_2017}. These RL policies directly map 2D laser scans to control actions, unlike classical methods that construct explicit maps from the laser scans. While showing promise, they often did not compare against classical approaches or failed to outperform them~\cite{xiao2022motion}. Some recent studies have benchmarked such RL-based approaches and found them superior in challenging problems with dense obstacles and narrow passages~\cite{xu_benchmarking_2023}. Instead of replacing the entire navigation stack with an RL policy, \textit{modular} approaches replace specific components like the local planner~\cite{chiang2019learning} or the exploration algorithm~\cite{stein_learning_2018} with RL, enabling better performance than classical baselines. However, these improvements were mainly observed in limited real settings. Most commercially deployed systems still primarily adopt classical stacks, owing to the lack of safety, interpretability, and generalization of RL-based methods~\cite{xiao2022motion, xu_benchmarking_2023}.  

\textbf{Visual Navigation.}
Visual navigation refers to problems where agents navigate to a goal based on visual observations. The additional input and task complexity pose challenges but enable agents to learn common strategies for navigating in similar environments (e.g., homes), where structural patterns emerge in visual data. Goals are typically specified as a point relative to the agent (termed \textit{pointgoal} navigation) or as an image of a particular object (\textit{objectgoal} or \textit{imagegoal}). RL is also commonly applied to vision-and-language navigation problems~\cite{anderson_vision-and-language_2018}, though very little work has demonstrated these capabilities on a real robot.  
Many works~\cite{zhu_target-driven_2017, kahn_self-supervised_2018} map visual observations to actions directly without mapping or planning modules. These end-to-end methods have achieved near-perfect results on pointgoal tasks in visually realistic simulations~\cite{wijmans_dd-ppo_2020}. However, training such policies is challenging due to the need for scene understanding, intelligent exploration, and episodic memory. Their applicability for real-world navigation remains unclear, as they have mostly been validated in limited real or lab settings.
Other works have investigated modular designs, e.g., using RL as a global exploration policy together with explicit mapping and local planning~\cite{chaplot_object_2020, gervet2023navigating}. They have outperformed both classical and end-to-end learning baselines on pointgoal and imagegoal tasks. 
However, some challenges with such modular approaches exist, such as dynamic obstacles, where end-to-end methods have shown promise~\cite{hoeller_learning_2021}. 

Despite the plethora of RL works on visual navigation, most are limited to simulation. While these simulators are typically constructed with real-world scans~\cite{anderson_vision-and-language_2018, savva2019habitat}, their transferability to the real world remains debatable. Some works reported poor transfer due to visual domain differences~\cite{gervet2023navigating}, while others found success through parameter tuning~\cite{kadian2020sim2real},  abstraction of dynamics~\cite{truong_rethinking_2023}, or employing only depth images rather than RGB-D~\cite{hoeller_learning_2021, truong_indoorsim--outdoorreal_2023}. 

\textbf{Off-road navigation.}
Navigating off-road presents additional challenges due to the dynamics and traversability of different terrains. 
Some methods tackled these challenges with model-based RL to learn predictive models of events or disengagements~\cite{kahn_badgr_2021}, or utilizing demonstration data with offline RL~\cite{shah_offline_2023}. Success has also been achieved in high-speed, off-road driving with model-based RL~\cite{williams2017information} and, recently, vision-based model-free RL~\cite{stachowicz_fastrlap_2023}. 

\begin{figure*}[t!]
    \centering
    \begin{subfigure}[c]{0.54\textwidth}
        \centering
        \includegraphics[width=0.95\textwidth]{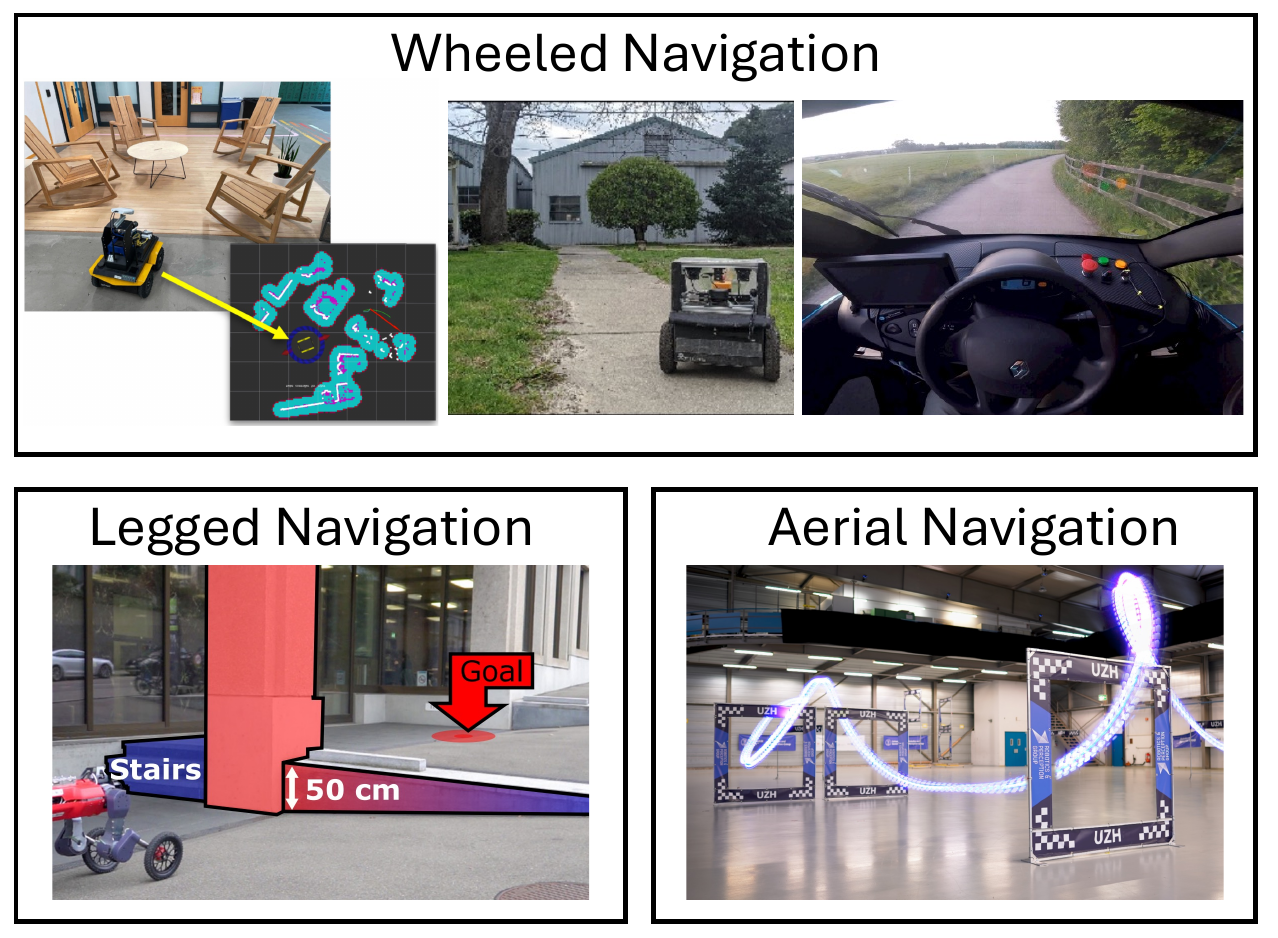}
    \end{subfigure}
    \begin{subfigure}[c]{0.45\textwidth}
        \centering 
        \begin{tabular}{P{1.3cm}|P{2.8cm}}
        \toprule 
        Wheeled & 
        \Level{1}{tai_virtual--real_2017},
        \Level{2}{xu_benchmarking_2023},
        \Level{3}{chiang2019learning},
        \Level{3}{stein_learning_2018},
        \Level{1}{zhu_target-driven_2017},
        \Level{3}{chaplot_object_2020},
        \Level{4}{gervet2023navigating},
        \Level{1}{kadian2020sim2real},
        \Level{4}{kahn_badgr_2021},
        \Level{3}{shah_offline_2023},
        \Level{1}{williams2017information},
        \Level{3}{stachowicz_fastrlap_2023},
        \Level{2}{kendall2019learning},
        \Level{4}{jang2024reinforcement}
          \\\hline 
        Legged & \Level{2}{rudin2022advanced}, \Level{2}{hoeller_learning_2021}, \Level{3}{truong_rethinking_2023}, \Level{3}{truong_indoorsim--outdoorreal_2023}, \Level{3}{sorokin2022learning}, \Level{2}{zhang2024resilient}, \Level{2}{hoeller2024anymal}, \Level{4}{lee2024learning}, \Level{3}{miki_learning_2024}, \Level{1}{xu2024dexterous}, \Level{3}{he2024agile} \\ \hline
        Aerial & \Level{3}{kaufmann2023champion}, \Level{3}{song2023reaching}, \Level{3}{sadeghi2017cad2rl}, \Level{3}{kang2019generalization}, \Level{3}{romero2023actor}\\ \bottomrule
        \end{tabular}
        \vspace{8pt}
    \end{subfigure}
    \caption{\textbf{Left:} An overview of the three navigation problems reviewed in Sec.~\ref{sec:navigation}, including wheeled navigation~\cite{xu_benchmarking_2023,kahn_badgr_2021,kendall2019learning}, legged navigation~\cite{lee2024learning}, and aerial navigation~\cite{song2023reaching}; \textbf{Right:} Navigation papers reviewed in Sec.~\ref{sec:navigation}. The color map indicates the levels of real-world success: \colorbox{L1}{\emph{\textcolor{black}{Limited Lab}}}, \colorbox{L2}{\emph{\textcolor{black}{Diverse Lab}}}, \colorbox{L3}{\emph{\textcolor{black}{Limited Real}}}, and \colorbox{L4}{\emph{\textcolor{black}{Diverse Real}}}.}\label{fig:navigation}
    \vspace{-6pt}
\end{figure*}

{\bf Autonomous Driving.} 
Autonomous driving extends wheeled navigation to full-size passenger vehicles operating at higher speeds in more complex and safety-critical environments. RL has achieved limited real-world success for autonomous driving~\cite{kiran2021deep} with a few examples under specific conditions. Kendall et al.~\cite{kendall2019learning} trained a lane-following policy by learning to maximize its progress before the safety driver intervenes. More recently, Jang et al.~\cite{jang2024reinforcement} trained a cruise control policy, where the policy command is wrapped by manually specified thresholds to ensure safety. They deployed their policy onto 100 vehicles to smooth traffic flow in a field test. Their work suggested a pragmatic approach to embed RL into self-driving stacks and showed its potential 
 benefits at the fleet level. 

\subsubsection{Legged Navigation}\label{sec:legged-nav} 
Legged navigation shares many challenges with wheeled navigation but also enables transversal of more complex terrains. Some have shown that robust visual-legged navigation policies can be learned with low-fidelity kinematic-only simulation for both indoors~\cite{hoeller_learning_2021,truong_rethinking_2023} and outdoors~\cite{truong_rethinking_2023,sorokin2022learning}. The policies thus focus on kinematic-level control while assuming effective low-level locomotion control during deployment. Truong et al.~\cite{truong_rethinking_2023} showed that this approach, in contrast to learning end-to-end policies with high-fidelity simulation, facilitates faster simulation and improves policy generalizability. With legged locomotion dynamics abstracted away, the approaches are similar to the wheeled case, with the main challenge being the visual domain gap. Unsupervised representation learning~\cite{hoeller_learning_2021} and pre-trained vision models~\cite{sorokin2022learning} have been used to facilitate robust visual policies. For outdoor scenes, Truong et al.~\cite{truong_indoorsim--outdoorreal_2023} zero-shot transferred policies trained in well-established indoor simulators to outdoors, using goal vector normalization and camera pitch randomization to bridge the indoor-to-outdoor domain gap. Sorokin et al.~\cite{sorokin2022learning} used a high-fidelity autonomous driving simulator and extracted visual features from a pre-trained semantic segmentation model for robust sim-to-real transfer to sidewalk navigation. 

While abstracting away low-level locomotion has advantages, it limits the system from fully utilizing the agile locomotion skills endowed by advanced locomotion controllers. Recent research has explored DRL frameworks integrating locomotion with navigation, achieving high-speed obstacle avoidance~\cite{he2024agile} and agile navigation over challenging terrains (e.g., stairs, gaps, and boxes)~\cite{rudin2022advanced,hoeller2024anymal} and through confined 3D space~\cite{miki_learning_2024,xu2024dexterous}. Particularly, Lee et al.~\cite{lee2024learning} demonstrated kilometer-scale navigation with a wheeled-legged robot in urban scenarios, overcoming challenging terrains and dynamic obstacles. The integrated policy network can be end-to-end, taking goal coordinates as input and outputting locomotion commands~\cite{rudin2022advanced, xu2024dexterous}. He et al.~\cite{he2024agile} further introduced a recovery policy coordinated using a learned reach-avoid value network. Alternatively, training efficiency can be improved with hierarchical architectures, where a high-level policy governs pre-trained low-level locomotion policies~\cite{zhang2024resilient,hoeller2024anymal,lee2024learning,miki_learning_2024}. 
Despite the potential of integrating locomotion with navigation, policy training could be costly and unstable due to the complex low-level dynamics together with the long-horizon nature and sparse rewards of the navigation tasks~\cite{rudin2022advanced,hoeller2024anymal}. Classical planning algorithms are often used for generating local waypoints to reduce the navigation horizon and synthesizing feasible paths to guide initial training~\cite{lee2024learning}. 

\subsubsection{Aerial Navigation} ompared to wheeled and legged robots, aerial vehicles such as quadrotors are more fragile, requiring higher robustness and safety in navigation policies. The weight and power constraints of quadrotors also limit the use of sophisticated sensors. Several works have explored DRL-based aerial navigation using low-cost monocular cameras~\cite{sadeghi2017cad2rl, kang2019generalization}. Sadeghi et al.~\cite{sadeghi2017cad2rl} leveraged visual domain randomization to achieve zero-shot sim-to-real transfer for indoor aerial navigation. Kang et al.~\cite{kang2019generalization} showed the values of 1) task-specific pre-training in simulation for learning generalizable visual representation and 2) the use of real-world data for learning accurate dynamics~\cite{kahn_self-supervised_2018}. Similar to quadruped navigation, DRL has been used to develop end-to-end navigation and locomotion policies for agile aerial navigation. Kaufmann et al.~\cite{kaufmann2023champion} achieved human champion-level performance in drone racing. A key recipe behind their success was augmenting simulation with data-driven residual models of the drone’s perception and dynamics. Their subsequent study~\cite{song2023reaching} showed that RL’s advantage over model-based methods lies in its ability to directly optimize the long-horizon racing task objective. However, DRL-based policies are still less robust than human pilots, limiting their operational conditions. Integrating actor-critic RL with differential MPC has shown promise in enhancing robustness~\cite{romero2023actor}.

\subsubsection{Trends and Challenges in Navigation}
RL has shown potential for various submodules of navigation systems, such as local planning~\cite{chiang2019learning, xie2017towards} and global exploration~\cite{stein_learning_2018, chaplot_object_2020, gervet2023navigating}, and for constructing end-to-end navigation solutions~\cite{xu_benchmarking_2023}. However, RL-based solutions for navigation lack the generalization, explainability, and safety guarantees of classical systems and thus have not seen widespread real-world deployment~\cite{xiao2022motion, xu_benchmarking_2023}. 

In visual navigation, model-free, end-to-end policies show promise for structured indoor environments like homes~\cite{wijmans2019dd}, while modular architectures boost performance without sacrificing guarantees and generalization~\cite{chaplot_object_2020, gervet2023navigating}. Striking the right balance between learned and classical modules remains an open challenge. Hybrid approaches may be promising, for example, leveraging implicit map-like representations learned by end-to-end approaches~\cite{wijmans2023emergence}, or using differentiable scene representations~\cite{rosinol2023nerf} to enable RL with algorithmic structure. 
RL-based vision-and-language navigation~\cite{anderson_vision-and-language_2018} is relatively under-explored in real-world settings but promising given the recent advances in vision-language models.

In legged navigation, abstracting away low-level dynamics has been shown to facilitate sim-to-real transfer for navigation~\cite{truong_rethinking_2023}. For agile legged and aerial navigation, where low-level complexity is unavoidable, jointly learning navigation and locomotion yields promising results~\cite{he2024agile, rudin2022advanced, hoeller2024anymal, kaufmann2023champion}. Yet, involving locomotion complicates the training of long-horizon navigation policies, which requires future developments to stabilize learning. 

Finally, learning navigation (collision avoidance, in particular) for safety-critical systems, like urban autonomous vehicles and drones, is challenging due to stringent robustness requirements in perception and control. These domains have seen fewer real-world successes as a result. Real-world data can help improve simulation fidelity for this purpose~\cite{kaufmann2023champion, song2023reaching, romero2023actor}, though establishing guarantees on their performance remains difficult.

\begin{summary}[Key Takeaways]
\begin{itemize}
    \item While end-to-end RL excels at visual navigation in simulation, most real-world successes deploy modular designs and learn components of the navigation stack. 
    \item Integrating RL into these modular architectures, e.g., for local planning or semantic exploration, is a promising avenue. 
    \item Recent work reasoning jointly about navigation and locomotion enables agile legged and aerial navigation, yet how to learn long-horizon navigation stably and efficiently with low-level control in the loop remains an open challenge.  
    \item Safety-critical applications like urban autonomous driving or outdoor drone flight have seen few real-world successes due to the higher requirements for robustness and the lack of explainability and generalization on the part of RL algorithms.
\end{itemize}
\vspace{-8pt}
\end{summary}
\subsection{Manipulation}\label{sec:manipulation}

\begin{figure*}[t!]
    \centering
    \begin{subfigure}[c]{\textwidth}
        \centering
        \includegraphics[width=0.95\textwidth]{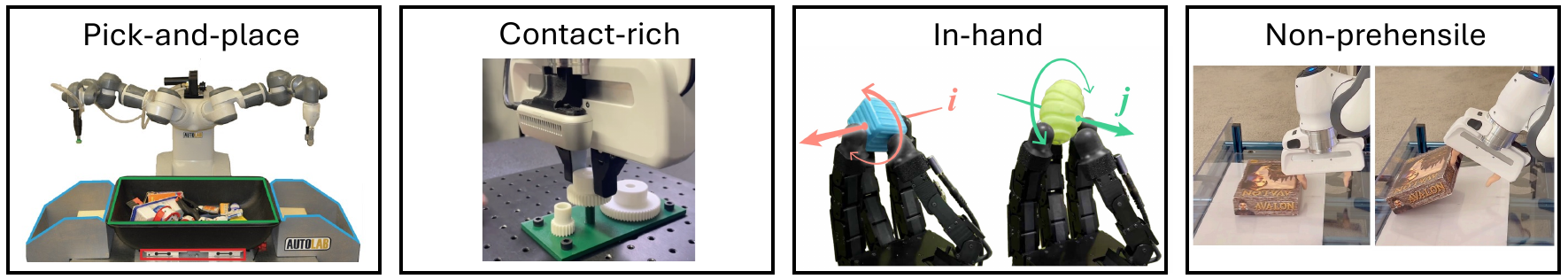}
    \end{subfigure}
    \begin{subfigure}[c]{\textwidth}
        \centering 
        \begin{tabular}{P{2.2cm}|P{3.1cm}|P{6.15cm}}
                \toprule 
        \multirow{2}{*}[-1em]{Pick-and-place} & 
        Grasping & 
           \Level{4}{mahler2019learning}, 
        \Level{2}{zeng2018learning}, 
        \Level{3}{kalashnikov_scalable_2018}, 
        \Level{2}{james_sim--real_2019},
        \Level{3}{wang2023robot}
        \\\cline{2-3}
        
         & End-to-end Pick-and-place & 
         \Level{1}{wu2023daydreamer},
         \Level{2}{levine2016end},
         \Level{3}{kalashnikov2022scaling},
        \Level{3}{chebotar_actionable_2021},
        \Level{3}{lee2021beyond},
        \Level{3}{walke_dont_2022},
        \Level{2}{ebert2018visual},
        \Level{3}{riedmiller_learning_2018},
        \Level{3}{zhu_ingredients_2020},
        \Level{1}{ma_vip_2022},
        \Level{1}{nair2020awac},
        \Level{3}{nasiriany2022augmenting},
        \Level{4}{chebotar_q-transformer_2023},
        \Level{1}{nair_visual_2018}
        \\ \hline
        \multirow{2}{*}[-1em]{Contact-rich} & 
         Assembly & 
        \Level{1}{johannink_residual_2019},
        \Level{1}{vecerik_leveraging_2018},
        \Level{4}{luo_robust_2021},
        \Level{4}{zhao_offline_2022},
        \Level{4}{tang2023industreal}
        \\\cline{2-3}
        & Articulated Objects & 
        \Level{1}{nair2020awac},
        \Level{3}{chebotar_closing_2019},
        \Level{2}{abbatematteo2024composable},
        \Level{3}{wu2022vat}
        \\\cline{2-3}

        & Deformable Objects &
        \Level{2}{matas_sim--real_2018},
        \Level{1}{wu_learning_2020},
        \Level{3}{avigal2022speedfolding},
        \Level{4}{wang_one_2023}\\\hline

        In-hand & --- & 
        \Level{3}{andrychowicz2020learning},
        \Level{3}{handa_dextreme_2023},
        \Level{3}{nagabandi_deep_2020},
        \Level{4}{qi2023general},
        \Level{4}{chen2023visual}, 
        \Level{2}{sievers_learning_2022} 
        \Level{3}{Pitz2024}\\ \hline
        
        Non-prehensile & --- &  
        \Level{2}{zeng2018learning},
        \Level{2}{ebert2018visual},
        \Level{3}{zhou2023learning}, 
        \Level{4}{zhou2023hacman},
        \Level{4}{cho2024corn}\\ 
        \bottomrule
        \end{tabular}
        \vspace{12pt}
    \end{subfigure}
    \caption{\textbf{Top:} An overview of the four manipulation problems reviewed in Sec.~\ref{sec:manipulation}, including pick-and-place~\cite{mahler2019learning}, contact-rich manipulation~\cite{tang2023industreal}, in-hand manipulation~\cite{qi2023general}, and non-prehensile manipulation~\cite{zhou2023learning}; \textbf{Bottom:} Manipulation papers reviewed in Sec.~\ref{sec:manipulation}. The color map indicates the levels of real-world success: \colorbox{L1}{\emph{\textcolor{black}{Limited Lab}}}, \colorbox{L2}{\emph{\textcolor{black}{Diverse Lab}}}, \colorbox{L3}{\emph{\textcolor{black}{Limited Real}}}, and \colorbox{L4}{\emph{\textcolor{black}{Diverse Real}}}.}\label{fig:manipulation}
\end{figure*}

Manipulation refers to an agent's control of its environment through selective contact~\cite{mason2018toward}. To perform useful work in the world, robots require manipulation capabilities such as pick-and-place, mechanical assembly, in-hand manipulation, non-prehensile manipulation, and beyond. Manipulation poses several challenges for both analytical and learning-based methods~\cite{kroemer_review_2019}, as the mechanics of contact are complex and difficult to model, and open-world manipulation requires strong generalization and fast online learning. RL is well-suited to these challenges, but manipulation poses fundamental difficulties for RL: large observation and action spaces make real-world exploration prohibitively time-consuming and unsafe; reward function design requires domain knowledge; tasks are often long-horizon; and instantaneous environment resets are usually unrealistic in real-world tasks. Despite these challenges, DRL has achieved notable successes in manipulation recently. 

In this subsection, we review progress in several manipulation capabilities enabled by DRL, following the outline from Mason's seminal review~\cite{mason2018toward}: \textbf{pick-and-place}, \textbf{contact-rich manipulation}, \textbf{in-hand manipulation}, and \textbf{non-prehensile manipulation}. See Figure~\ref{fig:manipulation} for an overview of the papers reviewed in this subsection. Note that this subsection focuses on stationary manipulators, and we defer mobile manipulation to Sec.~\ref{sec:moma}. 

\subsubsection{Pick-and-place}
\label{subsubsec:pickplace}
Picking and placing objects is a longstanding challenge in manipulation, requiring the ability to perceive objects, grasp them, determine appropriate placements, and generate collision-free motion. \textit{Structured} pick-and-place, in which the environment is engineered to reduce complexity and objects are known a priori, is well-understood and widely deployed in manufacturing contexts. Open-world, \textit{unstructured} pick-and-place---rearranging arbitrary objects in the wild---remains a challenge. In recent years, more traditional robotic approaches have seen success in industrial applications like fulfillment, employing machine learning for object detection and grasping but deferring control to analytical methods~\cite{mason2018toward}. While pick-and-place tasks serve as a common testbed for new RL algorithms~\cite{chebotar_actionable_2021, lee2021beyond, walke_dont_2022, ebert2018visual, riedmiller_learning_2018, wu2023daydreamer}, end-to-end RL methods still lack the ability to pick and place novel objects in the open world with generality. 
However, modular approaches, such as solving grasping with RL, have enabled some real-world successes. We will review RL-based solutions to the subproblem of grasping and then discuss end-to-end RL methods, omitting a discussion of motion generation for which RL is not commonly used.  

\paragraph{Grasping}
Grasping objects is a fundamental capability essential for pick-and-place and other downstream tasks, such as in-hand manipulation and assembly. 
Some of the first large-scale successes of DRL for manipulation were in grasping objects with unknown geometry and appearance~\cite{mahler2019learning}. Where analytical methods had achieved grasping of known objects using taxonomies and databases, these works leveraged thousands or millions of grasp attempts to learn grasping behaviors through interaction. Many works frame grasping as a bandit or classification problem, where the action space consists of discrete grasp candidates and the picking motion is executed open-loop~\cite{mahler2019learning, zeng2018learning}. These methods commonly employ sparse rewards that indicate success when an object is lifted and collect data in a self-supervisory manner.  
Similar systems have been reportedly integrated into fulfillment applications\footnote{See examples from Ambi Robotics (\url{https://t.ly/tSds_}) and Covariant (\url{https://t.ly/S5pnz}).} with diverse objects.
Closed-loop grasping---controlling the end-effector pose and/or fingers directly to achieve stable grasps---can be formulated as a sequential decision-making problem and solved with RL. While some successes have been seen~\cite{kalashnikov_scalable_2018, james_sim--real_2019, wang2023robot}, closed-loop grasping remains challenging due to the additional complexity of learning vision-based closed-loop control, and such systems have not seen the same level of real-world success as open-loop ones. In both closed- and open-loop grasping, while some works exclusively collect real-world data~\cite{zeng2018learning, kalashnikov_scalable_2018, wang2023robot}, the common recipe is to use simulation for data collection~\cite{mahler2019learning} or policy training~\cite{james_sim--real_2019}, often employing domain adaptation to ensure visual similarity between the simulator and real world. 



\paragraph{End-to-end Pick-and-place}
Learning general-purpose pick-and-place in the open world remains daunting for end-to-end RL, owing to the sheer variety of objects and tasks and the limited generalization of current algorithms. This variety also precludes the common sim-to-real recipe successful in other domains like grasping and in-hand manipulation, where tasks and objects can be enumerated during training. Nonetheless, some major milestones in end-to-end pick-and-place have been observed: Levine et al.~\cite{levine2016end} demonstrated the potential of deep visuomotor policies; Riedmiller et al.~\cite{riedmiller_learning_2018} demonstrated pick-and-place manipulation with a hierarchical policy trained in the real world; and  
Lee et al.~\cite{lee2021beyond} achieved stacking of diverse objects through sim-to-real transfer. 
Augmenting the action space with primitives~\cite{nasiriany2022augmenting} can help in reducing the task horizon and is a natural means to incorporate human engineering. 
Recent work leveraging large vision-language models shows promise in handling open-ended diverse objects and task objectives specified by natural language~\cite{chebotar_q-transformer_2023}. The potential of RL to solve this longstanding challenge is only now coming into focus with emerging large-scale robotic datasets and foundation models. 
Despite not yet achieving widespread success in real-world deployments, many important RL innovations have been demonstrated in pick-and-place problems, addressing challenges such as multi-task learning~\cite{kalashnikov2022scaling, chebotar_actionable_2021, nair_visual_2018}, sample efficiency~\cite{wu2023daydreamer}, defining and computing reward~\cite{zhu_ingredients_2020, ma_vip_2022}, resetting the environment~\cite{walke_dont_2022}, and utilizing human demonstrations or offline data~\cite{nair2020awac, lee2021beyond, chebotar_q-transformer_2023}. 

\subsubsection{Contact-rich Manipulation}
\label{subsubsec:contactrich}
While pick-and-place tasks are often assumed to be strictly kinematic, contact-rich tasks like mechanical assembly (e.g., peg insertion), interacting with articulated objects (e.g., opening doors), and manipulating deformable objects, require reasoning about dynamics and relaxing the rigid-body assumption of the objects.
We discuss several contact-rich tasks where RL has advanced the state of the art: assembly, articulated object manipulation, and deformable object manipulation. 

\paragraph{Assembly} Assembly tasks are crucial in manufacturing, and automating them is a longstanding challenge in robotics. Existing industrial solutions tend to rely on extensive engineering of the environment and robot motions, resulting in behaviors sensitive to small perturbations and costly to design. Assembly is challenging for RL due to the difficulty in controlling contact-rich interactions and the stringent requirements for accuracy and precision, coupled with the need to handle diverse object parts. While RL has not seen widespread deployment in industrial contexts, some notable successes have been observed in recent years. 
Many approaches employ sim-to-real transfer to achieve assembly~\cite{tang2023industreal}, though some train policies directly in the real world~\cite{johannink_residual_2019, vecerik_leveraging_2018, luo_robust_2021, zhao_offline_2022}, typically leveraging human demonstrations. Luo et al.~\cite{luo_robust_2021} notably compare against solutions provided by integrators and find their RL-based policies more robust to perturbation. 
A common strategy among approaches to assembly is using residual RL~\cite{johannink_residual_2019}, in which a residual policy is learned on top of a reference trajectory.
Most works assume that the object is already grasped before assembly. By contrast, Tang et al.~\cite{tang2023industreal} present a sim-to-real RL framework for the entire assembly pipeline, including object detection, grasping, and insertion, achieving diverse assembly tasks by leveraging recent advances in contact simulation and developing algorithmic advances for sim-to-real transfer.

\paragraph{Articulated Objects} 
\label{subsubsec:constrained}
    Some limited successes have been observed in constrained manipulation tasks like opening drawers. Most commonly, these tasks are used to demonstrate RL capabilities without dedicated efforts to realize practical deployment~\cite{nair2020awac, chebotar_closing_2019}. 
    Other works target this class of skills in particular~\cite{abbatematteo2024composable, wu2022vat} with limited success.

\paragraph{Deformable Objects}
Deformable objects, such as cloth, present additional challenges owing to the difficulty in accurately modeling soft materials. Tasks like cloth folding~\cite{matas_sim--real_2018, wu_learning_2020, avigal2022speedfolding} and assistive dressing~\cite{wang_one_2023} have thus received considerable attention in RL. These works often employ sim-to-real transfer~\cite{matas_sim--real_2018, avigal2022speedfolding, wang_one_2023}, and often simplify the tasks using primitives such as pick-and-place~\cite{wu_learning_2020} and flinging~\cite{avigal2022speedfolding}. 

In summary, open-world contact-rich manipulation inherits the challenges of unstructured pick-and-place (namely, generalization to novel objects and tasks) and the additional challenge of controlling contact-rich interactions. Nonetheless, some successes have been demonstrated in contact-rich tasks, particularly assembly and deformable objects, where tasks are predefined, objects are enumerable, and rigid grasps are usually assumed.

\subsubsection{In-hand Manipulation}
\label{subsubsec:inhand}
Humans exhibit many in-hand manipulation behaviors, re-orienting and re-positioning objects to facilitate downstream manipulation. 
Impressive strides in the development of these capabilities have been made with DRL in recent years, allowing agents to learn such complex in-hand manipulation behaviors with impressive generalization. Several works focused on re-orienting single objects to target configurations~\cite{andrychowicz2020learning, handa_dextreme_2023}, employing pose estimation modules trained in simulation or using proprioception alone~\cite{sievers_learning_2022}. Nagabandi et al.~\cite{nagabandi_deep_2020} similarly demonstrated rotating Baoding balls with model-based RL. While showing impressive dexterity, these works focus on manipulating known objects (e.g., a given cube) with low-dimensional observations. Recent methods leveraging vision and/or tactile data have demonstrated rotating arbitrary objects about arbitrary axes~\cite{qi2023general}, even against gravity~\cite{chen2023visual, Pitz2024}.
These approaches employ extensive domain randomization and typically leverage privileged information (e.g., object shape information, dynamic properties) and dense rewards when training in simulation. 
An open challenge is integrating these in-hand manipulation skills with other manipulation abilities (e.g., tool use), which require re-orientation to a target configuration suitable for a downstream task. 

\subsubsection{Non-prehensile Manipulation}
\label{subsubsec:nonprehensile}
Non-prehensile manipulation, namely moving objects without grasping, is crucial when objects are too large to be grasped, grasps are occluded, or in tool use. Object pushing abilities have long been demonstrated with RL~\cite{ebert2018visual}, and studied in connection to grasping~\cite{zeng2018learning, zhou2023learning}. Recently, general non-prehensile re-orientation of diverse objects has been enabled through sim-to-real transfer of RL policies~\cite{zhou2023hacman, cho2024corn}. Similar to in-hand manipulation, learning with privileged information (i.e., object geometry) before distilling a student policy is a common approach. Further work is warranted to integrate these skills with prehensile and in-hand behaviors and to develop \textit{extrinsic} dexterity, where the environment is used to facilitate manipulation. How to synthesize these capabilities for general-purpose, open-world manipulation remains an open question. 

\subsubsection{Trends and Open Challenges in Manipulation}
\label{subsubsec: trends_manipulation}
RL is beginning to achieve real-world success in various manipulation problems. Generally, RL has been more successful in domains where the space of tasks is more constrained---grasping, in-hand manipulation, and assembly---rather than less, e.g., end-to-end pick-and-place. These more constrained tasks allow for a priori reward design and zero-shot sim-to-real transfer, whereas open-world pick-and-place and contact-rich manipulation require generalizing to diverse objects and tasks. The limitations of physical simulation may also preclude scaling sim-to-real for contact-rich tasks. Differentiable simulation has shown promise for this challenge~\cite{lv_sam-rl_2023}. 
Open-world manipulation will require several advances, including scaling collections of simulated assets and tasks; few-shot sim-to-real~\cite{chebotar_closing_2019}; multi-task learning~\cite{kalashnikov2022scaling, nair_visual_2018}; learning autonomously in the real world~\cite{zhu_ingredients_2020, walke_dont_2022, wu2023daydreamer}; learning reward functions from examples~\cite{zhu_ingredients_2020} or human videos~\cite{ma_vip_2022}; and utilizing human demonstrations~\cite{vecerik_leveraging_2018}, offline data~\cite{nair2020awac} and foundation models~\cite{chebotar_q-transformer_2023}. Incorporating priors, such as symmetry~\cite{wang2023robot} and geometry~\cite{van2024geometric}, is promising for improving sample efficiency, generalization, and safety. 
Learning more complex behaviors, e.g. bimanual~\cite{chitnis2020efficient} or dynamic tasks like table tennis~\cite{buchler2022learning}, is another important avenue for future work~\cite{kroemer_review_2019,mason2018toward}. 

Additionally, action spaces are typically chosen by domain experts to match each problem at hand. Open-loop grasping tends to employ an abstraction of motion generation for reaching and closing the fingers, whereas closed-loop grasping, assembly, and end-to-end pick-and-place methods typically control the end-effector Cartesian pose or velocity. Most in-hand manipulation approaches control the fingers in configuration space, keeping the end-effector itself in a fixed position. Equipping one agent with these various manipulation abilities remains an important challenge for deploying capable manipulators in the real world. Moreover, many of these real-world successes are demonstrated on short-horizon tasks; further work is warranted to build agents that can reason over longer periods of time and compose learned abilities together to solve long-horizon tasks~\cite{kroemer_review_2019,nasiriany2022augmenting, abbatematteo2024composable, cheng2023league, funk2022learn2assemble}.

\begin{summary}[Key Takeaways]
\begin{itemize}
    \item RL solutions for manipulation are generally less mature than locomotion, with few deployments in the wild, yet there exist many impressive demonstrations in representative real-world conditions.

    \item Manipulation subproblems where tasks can be enumerated a priori---e.g., grasping, in-hand manipulation, assembly---allow for zero-shot sim-to-real transfer, facilitating many of the real-world successes. 

    \item Integrating manipulation subfields and connecting with task planning to build a generally competent manipulator remains an open challenge. 
\end{itemize}
\end{summary}

\subsection{Mobile Manipulation}\label{sec:moma}
\begin{figure*}[t!]
    \centering
    \begin{subfigure}[c]{0.95\textwidth}
        \centering
        \includegraphics[width=0.95\textwidth]{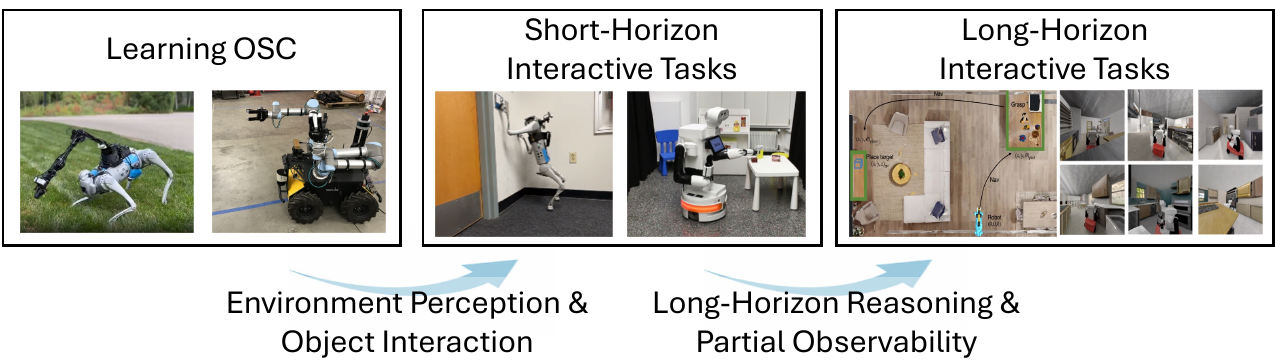}
    \end{subfigure}
    \begin{subfigure}[c]{\textwidth}    
        \begin{tabular}{P{2cm}|P{9.5cm}}
        \toprule 
        WBC & \Level{1}{wang2020learning},\Level{1}{ma2022combining},\Level{3}{fu2023deep},\Level{2}{fu2024humanplus} \\ \hline 
        
        Short-Horizon & \Level{1}{ji2022hierarchical},\Level{4}{liu2024visual},\Level{1}{sun2022fully},\Level{2}{jauhri2022robot},\Level{3}{ji2023dribblebot},\Level{3}{Hu-RSS-23},\Level{3}{honerkamp2023n},\Level{4}{uppal2024spin},\Level{3}{kumar2023cascaded},\Level{3}{yang2023harmonic},\Level{4}{xiong2024adaptive},\Level{3}{cheng2023legs}\\ \hline
        
        Long-Horizon & \Level{4}{yokoyama2023adaptive},\Level{4}{herzog2023deep},\Level{4}{wu2023m} \\ \bottomrule
        \end{tabular}
        \vspace{8pt}
    \end{subfigure}
    \caption{\textbf{Top:} An overview of the three MoMa challenges discussed in Sec.~\ref{sec:moma}, including whole-body control \cite{wang2020learning,fu2023deep} (WBC) and short- \cite{jauhri2022robot,cheng2023legs} and long-horizon \cite{yokoyama2023adaptive,wu2023m} interactive tasks; \textbf{Bottom:} MoMa papers reviewed in Sec.~\ref{sec:moma}.Color map indicates levels of real-world success: \colorbox{L1}{\textcolor{black}{\emph{Limited Lab}}}, \colorbox{L2}{\textcolor{black}{\emph{Diverse Lab}}}, \colorbox{L3}{\textcolor{black}{\emph{Limited Real}}}, and \colorbox{L4}{\textcolor{black}{\emph{Diverse Real}}}.}\label{fig:moma} 
    \vspace{-6pt}
\end{figure*}

Mobile manipulators are robotic agents combining mobility and manipulation competencies, unlocking applications in households, healthcare, and logistics.
Mobile manipulation (MoMa) problems present unique challenges requiring more than a simple concatenation of locomotion and manipulation, including the need to control and synchronize many degrees-of-freedom governing multiple body components (e.g., head, arm(s), and base/legs), strong partial observability and tasks with a natural long horizon. DRL has been applied to tackle various types of MoMa tasks, including 1) learning precise, real-time \textbf{whole-body control}; 2) learning object perception and interaction in \textbf{short-horizon interactive tasks}; and 3) high-level decision-making in \textbf{long-horizon interactive tasks}. 
In this section, we review works addressing these three problems summarized in Figure~\ref{fig:moma}.


\subsubsection{Learning Whole-Body Control}
The common goal in whole-body control (WBC) for mobile manipulators is to determine an action or sequence of actions for all degrees of freedom of the body to reach a desired configuration, possibly fulfilling additional constraints. 
Frequently, the desired configuration is specified as the desired position or pose of one or more of the links of the agent, e.g., the desired pose of the end-effector~\cite{wang2020learning, ma2022combining, fu2023deep,fu2024humanplus}. 
While there exist model-based analytical methods for whole-body control in advanced control theory literature~\cite{sentis2006whole}, DRL has been explored as a powerful alternative in situations where either the system dynamics are hard to model (e.g., leg-ground contact, slippery wheels, unknown manipulator dynamics), or when the inference-time computation is constrained\textemdash a frequent problem in MoMa tasks due to the robot embodiment's complexity. For example, Wang et al.~\cite{wang2020learning} and Fu et al.~\cite{fu2023deep} learned whole-body policies that enable a wheeled mobile manipulator and a quadruped with an arm to reach points in 3D space with their end-effector. Ma et al.~\cite{ma2022combining} learned a locomotion policy robust to random wrench perturbation and used an MPC planner to control the arm for point reaching. 

Typically, whole-body control problems focus on precise control of the end-effector without taking into account the agent's surroundings: the policy takes proprioceptive sensing as the observation and tries to minimize the difference to the desired configuration. 
Notably, recent works have explored how to integrate low-level whole-body control skills into hierarchical RL architectures~\cite{yokoyama2023adaptive, ji2022hierarchical,liu2024visual}, where the higher level perceives the surroundings and queries a low-level whole-body skill with the right desired pose as the goal. This extends the success of DRL in learning WBC to more complex interactive MoMa tasks.  

\subsubsection{Short-horizon Interactive Tasks}
Short-horizon interactive tasks often focus on learning specific sensorimotor skills that require no memory or planning capabilities.
Many works have explored applying DRL to these short-horizon tasks, including grasping~\cite{sun2022fully, jauhri2022robot, liu2024visual}, ball kicking~\cite{ji2023dribblebot, ji2022hierarchical}, collision-free target tracking~\cite{Hu-RSS-23, honerkamp2023n, uppal2024spin}, interactive navigation~\cite{kumar2023cascaded}, and door opening~\cite{yang2023harmonic, xiong2024adaptive}. 
Notably, 
Ji et al.~\cite{ji2022hierarchical} used hierarchical RL to learn soccer kicking skills, where a high-level policy generates the desired end-effector trajectory executed by a low-level policy. Hu et al.~\cite{Hu-RSS-23} improved the training efficiency by deriving a low-variance policy gradient update through action space decomposition.
Cheng et al.~\cite{cheng2023legs} learned separate skills for locomotion and manipulation on a quadruped in simulation and chained different skills using a behavior tree. Ji et al.~\cite{ji2023dribblebot} learned a whole-body dribbling policy in simulation, transferring it zero-shot to the real world using extensive domain randomization in visual input and simulation parameters. Liu et al.~\cite{liu2024visual} learned grasping policies via hierarchical RL and teacher-student distillation, where an image-based student policy is distilled from a state-based teacher policy. Interactive tasks require policies to make decisions based on sensor observations of their surroundings. Therefore, the policy usually takes in high-dimensional observations such as camera or lidar readings (Table~\ref{tab:taxonomy-mdp2}). Meanwhile, these tasks often involve hard-to-model dynamics such as contact forces or articulated object motion, making model-free RL an appealing alternative both to classical methods and to model-based RL (Table~\ref{tab:taxonomy-solution2}).

\subsubsection{Long-horizon Interactive Tasks}\label{sec:long-horizon-moma}
For a mobile manipulator to function in unstructured environments such as offices~\cite{herzog2023deep}, homes, or kitchens~\cite{wu2023m}, it needs to handle tasks with long horizons and strong partial observability. However, end-to-end RL struggles on long-horizon tasks due to the difficulty of exploring the state-action space to find successful strategies, requiring many samples to train. Partial observability is also challenging for DRL as it requires complex network architectures that can encode observation history (e.g., RNNs or LSTMs) or some other mechanism to aggregate observations and model the environment (e.g., mapping or 3D reconstruction). 
One possible way to mitigate this issue is to make use of expert demonstrations or simulation data to bootstrap the learning process. For instance, Herzog et al.~\cite{herzog2023deep} exploited simulation data and scripted policies to speed up the training process for off-policy RL in a waste sorting task.
Another promising direction is to take a divide-and-conquer approach by sequentially chaining short-horizon interactive skills through planning~\cite{wu2023m} or hierarchical RL~\cite{yokoyama2023adaptive}. Overall, solving long-horizon interactive tasks using DRL is an open challenge and under-explored area, but solving this type of task is necessary to create truly capable household and human-assistant robots.

\subsubsection{Trends and Open Challenges in Mobile Manipulation}
Thanks to the generalization of humanoids and other robot embodiments, and the advances in locomotion and stationary manipulation, DRL for MoMa is a growing field with increasing research attention. Based on our analysis, we infer some trends and open questions. First, compared to stationary manipulation, MoMa tasks have a significantly larger workspace, making safe real-world exploration challenging. As such, existing works mainly perform training in simulations where safety is not a concern (Table~\ref{tab:taxonomy-solution}). In the rare occurrences of real-world RL, strong domain knowledge, e.g., in the form of motion priors~\cite{xiong2024adaptive, sun2022fully} and/or demonstrations~\cite{xiong2024adaptive, herzog2023deep}, is used to enable safe and efficient exploration. 
Plus, MoMa's large workspace demands a more sophisticated form of memory and scene representation. Representations that work well for navigation often fail to capture the dynamic characters in manipulation.
While advances in sample efficiency, memory, and safe real-world RL promise new opportunities, scaling them to the open-worldness and vast workspaces inherent to MoMa remains challenging.

Second, mobile manipulators have very diverse morphologies compared to other types of robots, including wheeled robots with arms~\cite{herzog2023deep, Hu-RSS-23, yang2023harmonic, honerkamp2023n, wang2020learning, sun2022fully, jauhri2022robot, wu2023m, xiong2024adaptive}, quadrupeds with arms~\cite{ma2022combining, fu2023deep, liu2024visual, yokoyama2023adaptive}, humanoids~\cite{fu2024humanplus}, and even quadrupeds using their legs for both locomotion and manipulation, i.e., loco-manipulation~\cite{ji2022hierarchical, ji2023dribblebot, cheng2023legs, kumar2023cascaded}.
Each morphology brings unique challenges and opportunities. For example, wheeled mobile manipulators are easier to model and generally more kinematically stable, facilitating learning only for the manipulation component, while legged mobile manipulators can traverse uneven terrains but are harder to control, even for simple navigation phases. New research in both morphology-agnostic and morphology-specific RL methods is necessary for MoMa.

Third, perhaps due to the diverse morphologies, very diverse choices of action spaces are observed in the MoMa literature (Table~\ref{tab:taxonomy-mdp}), including direct joint control~\cite{Hu-RSS-23, ma2023learning, yang2023harmonic}, task-space control with classical model-based~\cite{honerkamp2023n, jauhri2022robot}, task-space control with learned low-level controllers~\cite{cheng2023legs, ji2022hierarchical, yokoyama2023adaptive}, and even factored actions that only controls a part of the embodiment~\cite{ma2022combining, honerkamp2023n}. Choosing the right action space is crucial for performance, as it affects the temporal abstraction levels and robot controllability. Yet, there is currently no principled way to select the appropriate action space for the diverse set of MoMa tasks. 
\begin{summary} [Key Takeaways]
    \begin{itemize}
        \item DRL has achieved initial success in mobile manipulation, in particular on short-horizon tasks, especially by leveraging training in simulation. 
        \item Defining a suitable action space is critical for RL in MoMa, especially given the diversity in the morphologies of existing MoMa systems.
        \item   The successes notwithstanding, existing methods are still insufficient for tackling multi-tasking, representing long-term memory, and performing safe exploration in the real world, providing opportunities for future improvements.
    \end{itemize}
    \vspace{-6pt}
\end{summary}


\subsection{Human-Robot Interaction}\label{sec:hri}
\begin{figure*}[t!]
    \centering
    \begin{subfigure}[c]{0.8\textwidth}
        \centering
        \includegraphics[width=0.95\textwidth]{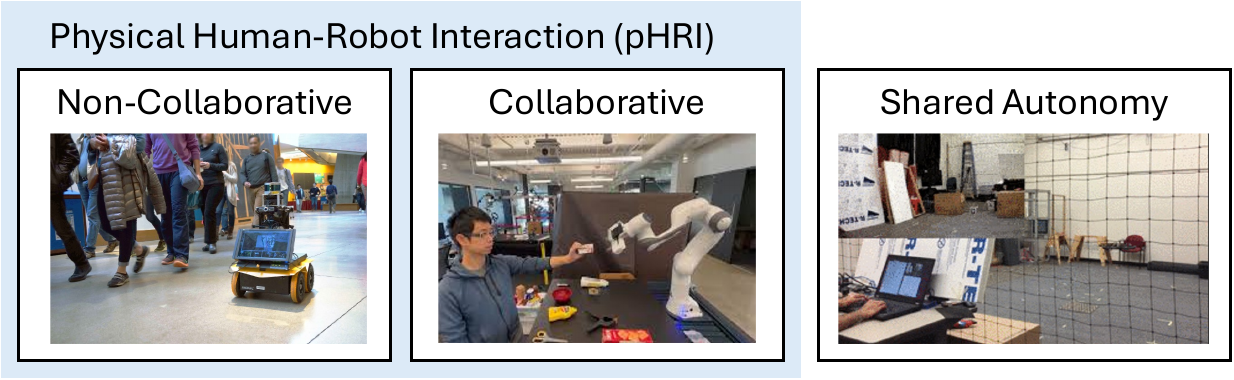}
        \vspace{-5pt}
    \end{subfigure}
    \begin{subfigure}[c]{0.8\textwidth}
        \centering 
        \begin{tabular}{P{3.5cm}|P{4.5cm}}
        \toprule 
        Collaborative pHRI & 
\Level{2}{christen2023synh2r}, \Level{1}{ghadirzadeh2020human}, \Level{2}{christen2023learning}, \Level{1}{dimeas2015comanip} \\ \hline 
        
        Non-collaborative pHRI & \Level{1}{chen2017socially},
\Level{2}{everett2021collision}, 
\Level{1}{liang2021crowd},
\Level{3}{hirose2024selfi}, \Level{1}{liu2023safe} \\ \hline
        
        Shared Autonomy & \Level{3}{nair_learning_2022},
\Level{1}{reddy2018shared},
\Level{0}{schaff2020residual} \\ \bottomrule
        \end{tabular}
        \vspace{10pt}
    \end{subfigure}
    \caption{\textbf{Top:} An overview of the three types of HRI tasks discussed in Sec.~\ref{sec:hri}, including collaborative \cite{christen2023synh2r} and non-collaborative \cite{chen2017socially} pHRI tasks, and shared autonomy \cite{reddy2018shared}; \textbf{Bottom:} Papers reviewed in Sec.~\ref{sec:hri}. The color map indicates the levels of real-world success: 
    \colorbox{L0}{\textcolor{black}{\emph{Sim Only}}}, \colorbox{L1}{\textcolor{black}{\emph{Limited Lab}}}, \colorbox{L2}{\textcolor{black}{\emph{Diverse Lab}}}, and \colorbox{L3}{\textcolor{black}{\emph{Limited Real}}}.}\label{fig:hri} 
    \vspace{-4pt}
\end{figure*}

In this subsection, we review works where DRL has been applied to human-robot interaction (HRI)\textemdash on robotic systems for use by or with humans. While HRI tasks can have varying objectives and involve robots with distinct morphology, the presence of humans introduces shared challenges, including safety, interpretability, and human modeling, that distinguish HRI from other robot problems not involving humans.
Notice that this section focuses on robotic systems with HRI \textit{competencies} (i.e., interact with humans \textit{during task execution}), whereas works that only involve humans during \textit{training} are out of the scope of this section. HRI tasks can be broadly classified into three main categories: \textbf{collaborative physical HRI (pHRI)}, where the robot and humans physically collaborate with a shared objective; \textbf{non-collaborative pHRI}, where the robot and humans share the same physical space but have distinct objectives; and \textbf{shared autonomy}, where humans act as teleoperators, and the robot autonomously interprets and executes the teleoperation command. In this section, we review works from these three categories. Figure~\ref{fig:hri} summarizes the papers reviewed. 

\subsubsection{Collaborative pHRI}
The most intuitive type of HRI arises when a robot and a human physically collaborate toward accomplishing a shared goal\textemdash a common theme for service robots that assist humans in household activities. For example, Ghadirzadeh et al.~\cite{ghadirzadeh2020human} tackled the collective packaging task, where recurrent Q-learning is combined with a behavior tree to minimize the packaging time of a human worker. Christen et al.~\cite{christen2023synh2r, christen2023learning} focused on object hand-over from a human to a robot, using RL to learn a simulated human hand-over policy and a robot policy to grasp the objects handed over by the human. Noticeably, existing works for collaborative pHRI share a similar procedure: learning a human model from pre-collected data to train a robot policy in simulation. This similarity is likely due to the high cost of collecting online interactions for collaborative tasks, which require continuous human attention and physical response to the robot's behavior. 



\subsubsection{Non-collaborative pHRI}
In non-collaborative pHRI tasks, a robot operates alongside humans in the same physical space but with different objectives. A representative example is social navigation， where a robot navigates through crowded environments. Chen et al. \cite{chen2017socially} trained a robot for social navigation in simulation, where a hand-crafted reward is used to encourage socially compliant behavior, and zero-shot transferred the policy to a real-world corridor. 
Everett et al.~\cite{everett2021collision} expanded on this work to incorporate human motion histories into decision-making by modeling the value network with an LSTM.
Liang et al.~\cite{liang2021crowd} developed a high-fidelity simulator of human motions to train navigation policies taking lidar scans as inputs, and demonstrated reliable sim-to-real transfer capabilities.
Hirose et al.~\cite{hirose2024selfi} learned navigation policies alongside humans in the real world. A residual Q-function is learned on top of an offline pre-trained Q-function to generate adaptive behavior on the fly. Unlike collaborative tasks, humans do not actively participate in the robot's activities in non-collaborative tasks, making it easier to hard-code human behaviors \cite{chen2017socially,everett2021collision,liang2021crowd} or train in the real world \cite{hirose2024selfi}, resulting in successful real-world implementations. Aside from social navigation, Liu et al.~\cite{liu2023safe} considered manipulation while avoiding collision with humans, where an action space transformation is conducted to ensure safe exploration in RL.

\subsubsection{Shared Autonomy}
Shared autonomy is an HRI paradigm that does not involve physical contact between humans and robots. Instead, the robot takes actions to complete tasks based on human instructions such as keyboard control or language commands. In this setting, RL can be used to learn a policy that conditions on human inputs and generates robot actions that optimize some external task rewards or constraints while aligning with the user instructions. 
For instance, Reddy et al.~\cite{reddy2018shared} tackled the quadrotor perching task, where a Q-function is learned based on task reward, and the robot chooses actions that are close to the user input and above a preset task value threshold.
Schaff et al.~\cite{schaff2020residual} formulated shared autonomy for simulated quadrotor control as a constrained optimization problem, where a residual RL policy is learned to minimally change the human input policy while satisfying a set of task-invariant constraints.
More recently, advances in NLP have opened up the possibility for shared autonomy through natural language instructions. For example, Nair et al.~\cite{nair_learning_2022} learned a language-conditioned policy for table-top manipulation using model-based RL on a pre-collected dataset with hand-labeled language instructions.

\subsubsection{Trends and Open Challenges in HRI}

Despite the importance of HRI for household robot applications, RL has seen fewer successes in HRI compared to other robotics domains like locomotion and manipulation. 
A primary challenge for applying RL to HRI problems is properly incorporating human or human-like priors into the training process, which can often be non-markovian, have limited rationality, and are often costly to collect. Existing works have primarily tackled this challenge in three ways. First, a straightforward approach is to train the policies directly in real-world environments alongside humans. 
However, this approach presents significant challenges to the sample complexity of the algorithm since collecting real-world interaction data is costly, especially when humans are actively involved. As such, works using this approach either focus on simple tasks \cite{dimeas2015comanip} or rely on pretraining to derive a good initial policy and reduce sample complexity \cite{hirose2024selfi}. Second, an alternative to avoid costly real-world learning is to learn a reasonable human model to simulate humans during training.
This approach is particularly appealing in domains where human actions are fairly easy to model, such as shared autonomy, where a human policy can be learned by imitating a set of human actors~\cite{schaff2020residual, reddy2018shared}.
In tasks where human actions are more complex, human models have been created using motion capture~\cite{ghadirzadeh2020human,liu2023safe}, crowd-sourcing~\cite{nair_learning_2022}, and RL~\cite{christen2023synh2r}. Third, when human behaviors are simple, human models can be directly hardcoded using domain knowledge~\cite{chen2017socially,everett2021collision,liang2021crowd}, and be incorporated either as parts of the simulation or as behavioral constraints. Although this approach is not scalable and inapplicable for many tasks, these simplified human models can serve as a useful source for pretraining to improve sample efficiency for real-world learning. 

Overall, two promising future directions emerge: first, developing safe and sample-efficient RL algorithms to enable direct real-world RL, possibly by leveraging known human behavior models; second, building high-fidelity human behavior simulation to bridge sim-to-real gaps for zero-shot sim-to-real transfer. 
Future advances in these directions promise to broaden the application of RL to HRI problems significantly.

\begin{summary} [Key Takeaways]
    \begin{itemize}
        \item Compared to other robotics domains, DRL has achieved limited success in HRI, especially on tasks that require the robot to collaborate with humans physically. 
        \item A key challenge for applying RL to HRI lies in collecting realistic interactive experiences with humans, which can, in principle, be obtained by either directly training in the real world or by building high-fidelity human models for simulations.
        \item Existing works have explored both approaches in simple tasks. However, whether and how we can scale up these approaches to more difficult tasks remains unclear.
    \end{itemize}
    \vspace{-6pt}
\end{summary}
\subsection{Multi-Robot Interaction}\label{subsec: multi-robot-interaction}
Multi-robot interaction is often solved as a MARL problem, which, in the most general case, is described using a partially observable stochastic game (POSG) with distinct reward functions and action and observation spaces, although most cooperative real-world problems model the problem as Decentralized POMDPs.
We highlight three real-world domains where DRL has been successfully applied to learn multi-robot interaction: \textbf{collision avoidance and navigation}, \textbf{multi-agent loco-manipulation}, and \textbf{robot soccer}. 

\subsubsection{Multi-Agent Collision Avoidance}
\label{subsubsec: multi_coll_avoid}

Chen et al.~\cite{chen2017decentralized} and Everett et al.~\cite{everett2018motion} model a Dec-MDP in which 
the policy takes the state vector consisting of positions, velocities, and radii of all the robots as input to predict the velocities for each robot.
The policy is preconditioned via finetuning using ORCA~\cite{van2011reciprocal}. 
The reward function is sparse, consisting of a goal-reaching reward and collision penalties. These works successfully developed collision-avoidance policies in simulation and showcased hardware results on aerial and ground robots. The multirotors used onboard sensors and controllers to execute maneuvers suggested by the policy. The ground robot, equipped with affordable onboard sensors (under $1000$ USD), was able to navigate through pedestrian traffic, effectively avoiding collisions despite imperfect perception and diverse pedestrian behaviors unseen during training. 

\begin{figure*}[t!]
    \centering
    \begin{subfigure}[c]{0.95\textwidth}
        \centering
        \includegraphics[width=0.85\textwidth]
        {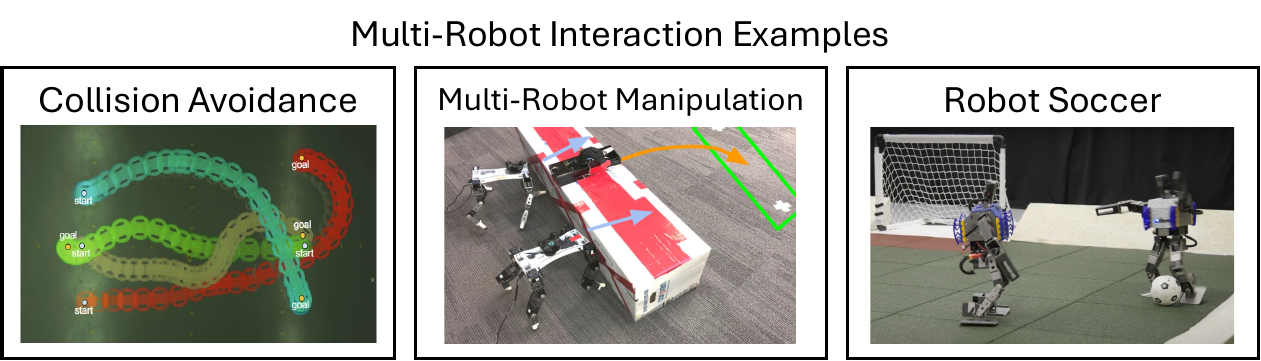}
        \vspace{-5pt}
    \end{subfigure}
    \begin{subfigure}[c]{0.95\textwidth}
        \centering 
        \begin{tabular}{P{5.2cm}|P{5.2cm}}
        \toprule 
        Multi-Robot Collision Avoidance & \Level{1}{chen2017decentralized}, \Level{1}{everett2018motion}, \Level{2}{fan2020distributed}, \Level{1}{han2022reinforcement}, \Level{1}{sartoretti2019primal}\\ \hline 
        
        
        Multi-Robot Loco-Manipulation & \Level{1}{nachum2019multi} \\ \hline
        
        Robot Soccer & \Level{2}{haarnoja2024learning}\\
        \bottomrule
        \end{tabular}
        \vspace{8pt}
    \end{subfigure}
    \caption{\textbf{Top:} An overview of the three representative multi-robot interaction domains reviewed in Sec.~\ref{subsec: multi-robot-interaction}, including multi-robot collision avoidance~\cite{fan2020distributed}, multi-robot manipulation via locomotion~\cite{nachum2019multi}, and robot soccer~\cite{haarnoja2024learning}; \textbf{Bottom:} Multi-robot interaction papers reviewed in Sec.~\ref{subsec: multi-robot-interaction}. See the caption of Fig.~\ref{fig:locomotion} for color map description.}\label{fig:marl} 
    \vspace{-8pt}
\end{figure*}

Other works~\cite{fan2020distributed, han2022reinforcement} have also modeled the problem as a Dec-MDP with the objective of time-to-goal minimization. These methods differ from the previous approaches in multiple respects. First, the policy takes raw lidar scans as input instead of the states of the other agents and thus does not depend on precise sensing and perception. Second, they do not precondition or finetune the policy using ORCA but instead employ curriculum learning and a dense reward function to facilitate training. Third, to deal with more complex multi-agent scenarios, it utilizes a hybrid controller to swap out the learned policy with a classical controller instead of restricting the other robots' motion via constant linear velocity models. 

Finally, Sartoretti et al.~\cite{sartoretti2019primal} used DRL to prevent agents from blocking each other in multi-agent pathfinding problems. A ``blocking penalty'' is applied when an agent reaches its goal but prevents another agent from doing the same. This strategy, combined with imitation learning and environment sampling, expedites convergence. The algorithm was tested on a small fleet of autonomous ground vehicles in a factory floor mock-up.


\subsubsection{Multi-Agent Loco-Manipulation}
\label{subsubsec: multi_manip}
We highlight a recent result~\cite{nachum2019multi} in multi-agent manipulation via locomotion (i.e., loco-manipulation). This involves multiple robots using movement to manipulate objects or interact with environments. Nachum et al.~\cite{nachum2019multi} focus on enabling multiple quadrupeds to perform complex tasks like manipulation and coordination using model-free RL. A significant challenge in applying RL to coordination or manipulation tasks with multiple legged robots is the complexity of interactions between agents or between agents and objects, which usually requires extensive real-world trial-and-error learning. To address this, this work employs a hierarchical sim2real approach demonstrating zero-shot sim-to-real transfer for object avoidance and targeted object pushing. Additionally, the work showcases a multi-agent scenario where two quadrupeds coordinate to move a heavy block to a specified location and orientation, illustrating the potential of using locomotion for coordinated multi-agent manipulation.

\subsubsection{Robot Soccer}
\label{subsubsec: multi_robot_soccer}
RL has also been successful in real physical soccer-playing robots in the RoboCup Standard Platform League.
Many of these works focus on training a policy for a single robot, which is then transferred to multiple robots. See Sec.~\ref{sec:locomotion} and Sec.~\ref{sec:moma} for discussions on these works focusing on single-robot competencies for robot soccer. A recent work~\cite{haarnoja2024learning} further applied RL to learn a variety of dynamic and complex movement skills like walking, turning, kicking, and rapid recovery from falls in \emph{1v1 robot soccer play}. The agents learn to apply skills appropriately via self-play and showcase sophisticated multi-agent competencies such as opponent interception.

\subsubsection{Trends and Open Challenges in Multi-Robot Interaction}
\label{subsubsec: multi_robot_open}

One of the most significant challenges in multi-agent systems is managing the complexity and scalability of the systems as the number of agents increases. This challenge is evident in multi-agent manipulation via locomotion and robot soccer, where the increase in team size exponentially escalates the complexity of the interactions.
The transition from controlled, simulated environments to unpredictable real-world conditions remains a formidable challenge. Although promising results have been shown in domains like collision avoidance, the variability in real-world dynamics, such as sensor inaccuracies, unexpected obstacles, and dynamic human interactions, often degrades system performance.
Next, while RL has provided impressive results in learning complex behaviors autonomously, integrating these learned behaviors with classical control methods is an increasingly popular area of research.
Finally, the ability of multi-robot systems to generalize across different tasks and environmental conditions presents a substantial opportunity for research. 

\vspace{-4pt}
\begin{summary}[Key Takeaways]
\begin{itemize}
    \item Current state-of-the-art in RL-based multi-robot interaction is limited to cooperative settings with identical reward functions, action spaces, and observation spaces.
    \item Predominantly, DRL in multi-robot settings is applied to collision avoidance among ground robots (as compared to manipulation via locomotion and robot soccer).
    \item Critical research areas moving forward include dealing with $(i)$ communication and networking between agents, $(ii)$ convergence and stability, $(iii)$ scalability, $(iv)$ general non-cooperative settings, $(v)$ different robot morphologies and applications.
\end{itemize}
\vspace{-8pt}
\end{summary}

\section{General Trends and Open Challenges}\label{sec:summary}

We conclude this survey by summarizing the patterns behind current real-world successes in robotics achieved with DRL and the characteristics of those less successful cases. Overall, more mature solutions (i.e., L3-4) have often followed the zero-shot sim-to-real transfer scheme (Table~\ref{tab:taxonomy-solution}), which works particularly well for locomotion and navigation. The dynamics involved in these competencies, especially terrestrial locomotion and navigation, are relatively stable and easy to simulate. Dense and shaped rewards, which simplify exploration and improve sample efficiency, have also been effective (Table~\ref{tab:taxonomy-mdp2}), leading to the predominant use of stable and robust model-free, on-policy algorithms in these domains (Table~\ref{tab:taxonomy-solution3}).
The sim-to-real scheme has been successful for manipulation problems in which dense reward functions can be designed a priori (e.g., grasping, assembly, in-hand, non-prehensile manipulation), but less so in tasks with more diversity (e.g., pick-and-place). 
The community has been striving to explore alternative solutions that do not require simulation (Table~\ref{tab:taxonomy-solution}) or reward shaping (Table~\ref{tab:taxonomy-solution2}) and adopt policy optimization algorithms with better sample efficiency (Table~\ref{tab:taxonomy-solution3}). 
Human demonstrations (Table~\ref{tab:taxonomy-solution2}) are effective for enabling real-world learning, particularly in manipulation tasks that are not prohibitively complex to demonstrate.
For competencies where both accurate simulation and real-world rollouts are prohibitive (e.g., HRI) or where stable, scalable RL algorithms are missing (e.g., multi-robot interaction), successful real-world examples are much sparser. In the remainder of this section, we identify several concrete open challenges that are opportunities for further extending DRL's applications, in particular for those currently less successful domains. 

\subsubsection*{Improving Stability and Sample-Efficiency in RL Algorithms.} While on-policy RL methods are often preferred due to their robustness to hyperparameters, collecting large amounts of on-policy data can be prohibitive, especially for real-world RL. Even in the predominant zero-shot sim-to-real setting, the sample efficiency of on-policy RL is problematic for tasks such as long-horizon mobile manipulation~\cite{herzog2023deep,wu2023m} and agile legged navigation~\cite{rudin2022advanced,hoeller2024anymal}, where the long task horizons, large operational spaces, sparse rewards, and complex contact dynamics hinder efficient exploration and stable learning. Sample efficiency can also be a crucial issue in problems with temporally extended action spaces~\cite{yang2020multi,gangapurwala2022rloc}. Fundamental algorithmic advances to develop RL algorithms that are at least as robust but more sample-efficient than on-policy methods are thus crucial for expanding RL's applications in robotics. An appealing direction is leveraging off-policy or offline samples to complement or replace on-policy exploration. However, off-policy and offline RL are often less stable due to the distributional shift between behavioral and learning policy experiences. Promising efforts have been made to derive scalable and more stable off-policy~\cite{kalashnikov_scalable_2018} and offline RL algorithms~\cite{chebotar_q-transformer_2023} for manipulation and MoMa~\cite{herzog2023deep}. Fine-tuning offline learned policies with online updates can further enhance performance in an efficient manner~\cite{smith2022legged,nair2020awac}. However, stable online fine-tuning is non-trivial, especially for value-based RL~\cite{uchendu2023jump,li2023residual}. Combining model-free and model-based approaches is another promising direction to derive sample-efficient RL algorithms~\cite{hansen2023td}. Lastly, these advances have primarily focused on single-robot problems. Multi-robot problems present greater challenges as the complexity of multi-robot interaction escalates exponentially with the number of robots. The scalability and stability of MARL remain open questions that hinder RL's application for multi-robot interaction. 

\subsubsection*{Real-World Learning.} In our analysis of RL for robot competencies (Sec.~\ref{sec:review}), real-world learning was often mentioned as one of the open challenges. A learning process carried out in the real world is crucial for robotic problems where the zero-shot sim-to-real transfer procedure is impractical due to the lack of high-fidelity simulation, such as open-world and contact-rich manipulation, lightweight quadrotor navigation, and physical HRI. Although some progress has been made, particularly for manipulation (Table~\ref{tab:taxonomy-solution}), successful real-world learning examples are much rarer than zero-shot sim-to-real transfer, presenting exciting opportunities for future research. Two main issues need to be addressed for real-world RL learning. The first issue is \emph{how to collect many useful experiences in a safe manner?} In domains where oracle policies, like humans~\cite{chebotar_q-transformer_2023} and scripts~\cite{herzog2023deep}, are available, demonstrations can be collected for offline learning. However, offline RL faces challenges such as distributional shifts, and the demonstration data can be suboptimal and costly to collect for human experts. Real-world rollouts require automatic resets~\cite{zhu_ingredients_2020, walke_dont_2022, smith2023demonstrating} and safe exploration mechanisms~\cite{xiong2024adaptive} to minimize human effort and ensure safety. Such mechanisms are still missing in most problem domains and present an opportunity for future development, especially for safety-critical applications~\cite{kendall2019learning,kang2019generalization}. To date, human-in-the-loop learning (for resets and safety) is currently the only alternative~\cite{kendall2019learning}, leaving automated real-world learning a desirable future capability. In addition to procedural and algorithmic improvements, safe real-world exploration may also be facilitated through hardware advances, such as adaptive and less fragile hardware and mechanisms that ensure safety passively~\cite{jeong2023bariflex}. The second issue is \textit{how do we accelerate training to require fewer experiences?} A promising avenue is to explore what modules can be updated with real-world samples and how. Instead of updating the entire policy with model-free RL, some solutions explore adapting vision encoders~\cite{loquercio2023learning} or learning (residual) dynamics models~\cite{kaufmann2023champion,kang2019generalization,hanna2021grounded} from real-world samples. These alternatives improve efficiency; we predict future successful real-world training procedures exploring alternative combinations of frozen-trainable modules.

\subsubsection*{Learning for Long-Horizon Robotic Tasks.} 
Long-horizon tasks pose a fundamental challenge to RL algorithms, requiring directed exploration and temporal credit assignment over long stretches of time. Many such real-world tasks require integrating diverse abilities. By contrast, the vast majority of the RL successes we have reviewed are in short-horizon problems, e.g., controlling a quadruped to walk at a given velocity or controlling a manipulator to rotate an object in hand. 
A promising avenue for solving long-horizon tasks is learning \textit{skills} and composing them, enabling compositional generalization. This approach has seen success in navigation~\cite{hoeller2024anymal,lee2024learning}, manipulation~\cite{kroemer_review_2019, chebotar_actionable_2021, abbatematteo2024composable, cheng2023league}, and MoMa~\cite{yokoyama2023adaptive,wu2023m}. 
A critical question for future work is: \textit{what skills should the robot learn?}. While some successes have been achieved with manually specified skills and reward functions~\cite{yang2020multi,zhuang2023robot,hoeller2024anymal, nasiriany2022augmenting, cheng2023league, kalashnikov2022scaling}, these approaches heavily rely on domain knowledge. Some efforts have been made to explore unified reward designs for learning multi-skill locomotion policies~\cite{fu2022minimizing,vollenweider2023advanced,cheng2023parkour}. Formulating skill learning as goal-conditioned~\cite{nair_visual_2018} or unsupervised RL~\cite{eysenbach2018diversity,schwarke2023curiosity} is promising for more general problems. A second critical question is: \textit{how should these skills be combined to solve long-horizon tasks?} Various designs have been explored, including hierarchical RL~\cite{yang2020multi,yokoyama2023adaptive}, end-to-end training~\cite{nasiriany2022augmenting, cheng2023parkour}, and planning~\cite{abbatematteo2024composable, cheng2023league, wu2023m}. This question will also be central to integrating various competencies toward general-purpose robots; recent advances along this line have opened up exciting possibilities, including wheel-legged navigation~\cite{lee2024learning} and loco-manipulation~\cite{vollenweider2023advanced,ji2022hierarchical, ji2023dribblebot, cheng2023legs, kumar2023cascaded}.

\subsubsection*{Designing Principled Approaches for RL Systems.} For each robotic task, an RL practitioner must choose among the many alternatives that will define its RL system, both in the problem formulation and solution space (see Table~\ref{tab:taxonomy-mdp}--\ref{tab:taxonomy-solution4}). Many of these choices are made based on expert knowledge and heuristics, which are not necessarily optimal and can even harm performance~\cite{xie2021dynamics, martin-martin_variable_2019}. Principled approaches for RL system design, relying less on heuristics and manual efforts, will be essential in the future for scalable development and deployment, especially for open-world tasks. Here, we note some particularly important examples. First, many real-world successes have been achieved with dense and shaped rewards designed with heavy engineering efforts, particularly in locomotion and navigation (Table~\ref{tab:taxonomy-mdp2}). Efforts are being made to explore principled reward designs for specific competencies~\cite{fu2022minimizing,vollenweider2023advanced,cheng2023parkour} and more general problems using goal-conditioned~\cite{nair_visual_2018} or unsupervised RL~\cite{eysenbach2018diversity,schwarke2023curiosity}. Second, various action spaces are used, particularly for manipulation and MoMa (Table~\ref{tab:taxonomy-mdp}). The action space choices affect the temporal abstraction levels and robustness of the RL policies. 
Some studies have attempted to benchmark different action spaces~\cite{kaufmann2022benchmark,truong_rethinking_2023,martin-martin_variable_2019}, but such principled studies and guidelines are still lacking for many problems. Another related design choice is the integration of RL with classical planning and control modules. The different levels of integration result in different action spaces for the RL policies (i.e., low-, mid-, and high-level). The effectiveness of end-to-end versus hybrid modular solutions varies by problem~\cite{gervet2023navigating,truong_rethinking_2023,he2024agile,song2023reaching,xia21relmogen}. Neither approach is universally superior. There are many other dimensions that require such principled investigations, which are crucial for advancing DRL's real-world success, in addition to exploring new frontiers in algorithms and applications. 

\subsubsection*{Benchmarking Real-World Success.} In this survey, we classify papers into six levels of real-world success to assess the maturity of DRL-based solutions. However, precisely determining these levels can be challenging since the only source of information is the experimental results reported by the authors, but the varying testing conditions and evaluation metrics make direct comparison difficult. This highlights the need for \emph{standard evaluation protocols and benchmarks for real-world performance}. While widely adopted, low-cost hardware, as seen with quadrupeds, is helpful by enabling standardized experimental platforms, it is not sufficient alone. Test environments and tasks must also resemble real-world conditions and, more importantly, be \emph{reproducible}. Multiple real-world benchmarks have been established, including those for manipulation~\cite{luo2024fmb,heo2023furniturebench} and domestic service robots~\cite{van2011robocup}. However, when it comes to complex open-world problems, the evaluation procedure must also scale up to be realistic and informative~\cite{li2024evaluating}. 
Overall, developing scalable evaluation protocols and benchmarks remains an exciting open research direction for many problems.


\subsubsection*{Leveraging Foundation Models.} Lastly, recent advances in large-scale robot dataset \cite{padalkar2023open,khazatsky2024droid} and robot foundation models~\cite{firoozi2023foundation,hu2023toward} present exciting open opportunities for RL successes in the real world. Foundation models have demonstrated impressive generalization capabilities across domains for reasoning and decision-making tasks~\cite{yang2023foundation}, showing promise for addressing several of the aforementioned challenges of DRL for robotics. For instance, the recently introduced DrEureka~\cite{ma2024dreureka} algorithm leverages large language models (LLMs) to automate reward design and domain randomization configuration for sim-to-real transfer without manual tuning. 
In addition, LLMs and vision-language models (VLMs) open up new opportunities to create language-conditioned RL policies for novel applications. We refer readers to existing surveys for detailed discussions on the opportunities foundation models offer in general~\cite{firoozi2023foundation,hu2023toward}, but we anticipate an increased integration of foundation models into RL solutions for real-world robotic tasks.  

\section{Conclusion}
Deep reinforcement learning has recently played an important role in the development of many robotic capabilities, leading to many real-world successes. Here, we have reviewed and categorized these successes, delineating them based on the specific robotic competency, problem formulation, and solution approach. Our analysis across these axes has revealed general trends and important avenues for future work, including algorithmic and procedural improvements, ingredients for real-world learning, and holistic approaches toward synthesizing all the competencies discussed herein. 
Harnessing RL's power to produce capable real-world robotic systems will require solving fundamental challenges and innovations in its application; nonetheless, we expect that RL will continue to play a central role in the development of generally intelligent robots. 

\vspace{-6pt}
\section*{Acknowledgements}
We thank Pieter Abbeel, Yuchen Cui, Shivin Dass, George Konidaris, Jan Peters, Eric Rosen, Koushil Sreenath, Eugene Vinitsky, and Zhaoming Xie for their feedback on the manuscript. We also thank Google DeepMind for permission to use representative images from their work on robot soccer. A portion of this work has taken place in the Learning Agents Research Group (LARG) at the Artificial Intelligence Laboratory at the University of Texas at Austin. LARG research is supported in part by the National Science Foundation (FAIN-2019844, NRT-2125858), the Office of Naval Research (N00014-18-2243), Army Research Office (E2061621), Bosch, Lockheed Martin, and Good Systems, a research grand challenge
at the University of Texas at Austin. The views and conclusions contained in this document are those of the authors alone. Peter Stone serves as the Chief Scientist of Sony AI and receives financial compensation for this work.  The terms of this arrangement have been reviewed and approved by the University of Texas at Austin in accordance with its policy on objectivity in research.

\bibliographystyle{ar-style3}
\bibliography{ref-v2}

\newpage
\begin{appendices}
\section{Term Definition}
As presented in Sec.~\ref{ss:taxonomy} of the main article, we classify the literature based on a taxonomy consisting of four axes: \textbf{robot competencies learned with DRL}, \textbf{problem formulation}, \textbf{solution approach}, and \textbf{the level of real-world success}. 
In this section, we provide a detailed definition and discussion of the elements along the \textbf{problem formulation} and \textbf{solution approach} axes. 

\subsection{Problem Formulation}
As discussed in Sec.~\ref{ss:taxonomy}, we categorize the papers based on the following elements of the problem formulation: 1) \emph{Action space}: whether the actions are \emph{low-level} (i.e., joint or motor commands), \emph{mid-level} (i.e., task-space commands), or \emph{high-level} (i.e., temporally extended task-space commands or subroutines); 2) \emph{Observation space}: whether the observations are \emph{high-dimensional} sensor inputs (e.g., images and/or LiDAR scans) or estimated \emph{low-dimensional} state vectors; 3) \emph{Reward function}: whether the reward signals are \emph{sparse} or \emph{dense}. This subsection provides detailed definitions and a discussion of these terms. 

\subsubsection{Action Space}~\\
\noindent\textbf{Low-level Actions}: We define low-level actions as those that directly operate in the robot's joint space, such as controlling torques of individual joints in a robot arm or velocities of individual wheels in a mobile robot. A low-level action space requires minimal domain knowledge and allows the policy to have fine-grained control over the robot's behavior. However, performing learning in low-level action spaces presents several challenges: 1) exploration with low-level actions is difficult, as random joint actions often result in trivial behaviors; 2) the action space scales linearly with the robot's degrees of freedom, often resulting in high-dimensional action spaces; and 3) joints are often controlled at a high frequency, resulting in extended task horizons and inference-time constraints.\\

\noindent\textbf{Mid-level Actions}: Mid-level actions control the robot in its workspace, such as adjusting the end-effector pose of a robot arm or controlling the velocity of the center of mass of a mobile robot. Once the policies generate these mid-level actions, they are often executed by an external controller, such as an inverse kinematics (IK) controller, to produce the joint-level torques. As such, operating in a mid-level action space requires domain knowledge to define an appropriate operational space and to design and implement the external controller effectively. When chosen correctly, mid-level action spaces strike a balance between incorporating domain knowledge and maintaining generality for various tasks. This approach is a popular choice in many RL applications for robotics, as it leverages specific expertise while allowing flexibility across different robotic functions.\\

\noindent\textbf{High-level Actions}: High-level actions control the robot through temporally extended ``skills" that can realize certain short-horizon behaviors, such as ``grasping certain objects" or ``moving to certain rooms." A well-designed high-level action space can greatly enhance the efficiency of the RL agent's exploration by drastically shortening the task horizon and ensuring that the robot performs task-relevant actions most of the time. However, designing an appropriate set of skills for the high-level action space is a complex problem, often requiring each skill to be formulated as an RL problem in itself. Additionally, these skills may not always be transferable across tasks, posing challenges to their scalability.

\subsubsection{Observation Space}~\\
\noindent\textbf{Low-dimensional Observations}: The robot's observations are represented as a compact, low-dimensional vector, which can include proprioceptive information, object locations, and task information.\\

\noindent\textbf{High-dimensional Observations}: The robot's observations include high-dimensional sensor data for exteroceptive information, which can be in the form of lidar readings, camera images, and/or point clouds.

\subsubsection{Reward Function}~\\
\noindent\textbf{Sparse Reward}: A sparse reward signal means the agent receives trivial reward signals for most of the potential transitions in a (PO)MDP and only receives non-trivial reward signals sparsely. One natural way of defining a sparse reward for a task is to have +1 for all transitions into a success termination state, -1 for all transitions into a failure termination state, and 0 for any other transitions.\\

\noindent\textbf{Dense Reward}: A dense reward means the reward signal is abundant, providing rich feedback to the agent. In certain tasks, such as locomotion, the reward is naturally dense (e.g., the error between the robot's current forward velocity and the instructed velocity). In other scenarios where the task reward is inherently sparse, such as navigation tasks, a dense reward can be defined by adding shaping components to the sparse reward (e.g., the distance between the robot and the navigation target). Such kind of shaped and dense rewards are often used to facilitate learning efficiency, especially for long-horizon tasks. 

\subsection{Solution Approach}
As introduced in Sec.~\ref{ss:taxonomy}, we classify the solution approach from the following perspectives: 1) \emph{Simulator usage}: whether and how simulators are used, categorized into \emph{zero-shot}, \emph{few-shot sim-to-real transfer}, or directly learning offline or in the real world \emph{without simulators}; 2) \emph{Model learning}: whether (a part of) the transition dynamics model is learned from robot data; 3) \emph{Expert usage}: whether expert (e.g., human or oracle policy) data are used to facilitate learning; 4) \emph{Policy optimization}: the policy optimization algorithm adopted, including \emph{planning} or \emph{offline}, \emph{off-policy}, or \emph{on-policy RL}; 5) \emph{Policy/Model Representation}: Classes of neural network architectures used to represent the policy or dynamics model, including \emph{MLP}, \emph{CNN}, \emph{RNN}, and \emph{Transformer}. This subsection provides detailed definitions of these terms.

\subsubsection{Simulator Usage}~\\
\noindent\textbf{Zero-shot sim2real}:
The training is performed entirely in a simulator, where the trained policy is deployed directly in the real world without additional learning.\\

\noindent\textbf{Few-shot sim2real}:
The robot is pre-trained in the simulator and fine-tuned in the real world with limited additional real-world interactions.\\

\noindent\textbf{No Simulator}: The training is conducted in the real world without using a simulator. 

\subsubsection{Model Learning} RL algorithms can be broadly classified into two categories: model-free RL and model-based RL, based on whether they learn a dynamics model. In model-free RL algorithms, such as PPO and SAC, the robot directly learns a policy or value function without explicitly modeling the environment's dynamics. Model-free RL is often easier to implement and superior when learning a good policy is simpler than learning a good model. In contrast, model-based RL algorithms, such as TD-MPC and Dreamer, involve the robot learning a world model that can predict the consequences of its actions. This world model can be used either for model-based planning and control or for generating experiences for a model-free RL agent, potentially increasing the agent's sample efficiency. Instead of learning the full dynamics, some methods learn a residual or a part of the dynamics model (e.g., actuator dynamics model) to complement the simulation for model-free RL, which we also mark as involving model learning in our categorization. 

\subsubsection{Expert Usage}
\textit{Tabula rasa} RL begins with random initialization, training entirely through trial and error. However, in robotics, it is sometimes possible to utilize an external expert to expedite the learning process. These experts may include human demonstrations, trajectory planners, oracle actions, and so on. In this survey, we classify all works that utilize an external expert, either offline or online, to facilitate learning as works ``with experts'', which gives them an advantage over methods that do not assume access to experts.

\subsubsection{Policy Optimization}~\\
\noindent\textbf{Planning}:
The robot’s policy is derived by solving an optimal control problem online using a learned world model. Representative algorithms include A* and MPPI (Model Predictive Path Integral).\\

\noindent\textbf{Offline}:
The robot does not interact with the environment during learning. Instead, it learns a policy and, optionally, a value function directly from offline data. Representative algorithms include CQL (Conservative Q-Learning) and DT (Decision Transformer).\\

\noindent\textbf{On-policy}:
The robot interacts with the environment during learning and only updates the policy with transitions collected by the current policy. Representative algorithms include PPO (Proximal Policy Optimization) and TRPO (Trust Region Policy Optimization).\\

\noindent\textbf{Off-policy}:
The robot interacts with the environment during learning and updates the policy with transitions collected by both the current policy and other/previous policies. Representative algorithms include SAC (Soft Actor-Critic) and DQN (Deep Q-Network).

\subsubsection{Policy/Model Representation}~\\
\noindent\textbf{MLP Only}:
Multi-layer Perceptron (MLP) models take 1D vector inputs and consist solely of fully connected layers. They are widely used for processing low-dimensional observations.\\

\noindent\textbf{CNN}:
Convolutional Neural Networks (CNNs) are a specialized type of MLP that preserves local spatial coherence, initially designed for image processing. Later works have extended CNNs to process 1D data like lidar readings and observation memory, as well as 3D data such as point clouds.\\

\noindent\textbf{RNN}:
Recurrent Neural Networks (RNNs), including LSTM and GRU, are bidirectional neural networks with internal memory. They are suitable for processing time-series data, such as trajectories over time.\\

\noindent\textbf{Transformer}:
Transformers take a sequence of vectors (tokens) as input and use multi-head self-attention to generate outputs. Recently, transformers have been widely used to process time-series data, natural language instructions, and visual information. They have also proven powerful in fusing tokenized multi-modal information.

\section{Additional Tables}
This section contains tables that present the complete categorization of the reviewed papers along all four axes of our taxonomy. Tables~\ref{tab:taxonomy-mdp}-\ref{tab:taxonomy-mdp2} categorize the papers based on problem formulation for each robot competency, while Tables~\ref{tab:taxonomy-solution}-\ref{tab:taxonomy-solution4} categorize the papers based on solution approach for each robot competency. As in the tables in the main article, the color map indicates the levels of real-world success: \colorbox{L0}{\emph{\textcolor{black}{Sim Only}}}, \colorbox{L1}{\emph{\textcolor{black}{Limited Lab}}}, \colorbox{L2}{\emph{\textcolor{black}{Diverse Lab}}}, \colorbox{L3}{\emph{\textcolor{black}{Limited Real}}}, and \colorbox{L4}{\emph{\textcolor{black}{Diverse Real}}}. 

In these tables, we add a superscript $^*$ to papers that appear in multiple columns, which means they adopt two different elements jointly (e.g., a hierarchical policy that outputs both low-level and mid-level actions, a policy network consists of both CNN and RNN).  

\begin{table}[ht]\fontsize{8pt}{8pt}\selectfont
\centering
\begin{tabular}{R{1.7cm}P{3.1cm}P{3.1cm}P{3.1cm}}
\toprule
 & \multicolumn{3}{c}{Action Space} \\ \cmidrule(lr){2-4}
Application & Low-Level & Mid-Level & High-Level \\ \hline
Locomotion & 
\Level{2}{kumar2022adapting}, \Level{1}{tan2018sim}, \Level{1}{hwangbo2019learning}, \Level{2}{feng2023genloco}, \Level{2}{lee2019robust}$^*$, \Level{3}{yang2020multi}$^*$, \Level{4}{kumar2021rma}, \Level{4}{gangapurwala2022rloc}$^*$, \Level{3}{choi2023learning}, \Level{4}{nahrendra2023dreamwaq}, \Level{1}{escontrela2022adversarial}, \Level{1}{ma2023learning}, \Level{3}{fu2022minimizing}, \Level{4}{loquercio2023learning}, \Level{4}{agarwal2023legged}, \Level{4}{yang2023neural}, \Level{4}{jenelten2024dtc}, \Level{3}{smith2022legged}, \Level{3}{cheng2023parkour}, \Level{3}{zhuang2023robot}, \Level{3}{vollenweider2023advanced}, \Level{3}{margolis2023walk}, \Level{3}{smith2023demonstrating}, \Level{1}{wu2023daydreamer}, \Level{1}{siekmann2020learning}, \Level{1}{hanna2021grounded}, \Level{1}{siekmann2021sim}, \Level{3}{li2021reinforcement}, \Level{3}{siekmann2021blind}, \Level{2}{duan2023learning}, \Level{3}{radosavovic2023real}, \Level{3}{li_reinforcement_2024}, \Level{1}{hwangbo2017control}, \Level{2}{molchanov2019sim}, \Level{2}{eschmann2024learning} & 
\Level{2}{lee2019robust}$^*$, \Level{3}{yang2020multi}$^*$, \Level{4}{lee2020learning}, \Level{4}{miki2022learning}, \Level{3}{yang2023cajun}, \Level{3}{kaufmann2022benchmark}, \Level{2}{zhang2023hover} & 
\Level{4}{gangapurwala2022rloc}$^*$, \Level{3}{castillo2022reinforcement}\\ \hline
Navigation & 
\Level{2}{rudin2022advanced}, 
\Level{1}{williams2017information},
\Level{2}{hoeller2024anymal}$^*$, 
\Level{4}{lee2024learning}$^*$, 
\Level{1}{xu2024dexterous}, 
\Level{3}{he2024agile}, 

& 
\Level{3}{kaufmann2023champion}, \Level{3}{song2023reaching}, 
\Level{1}{tai_virtual--real_2017}, 
\Level{2}{xu_benchmarking_2023}, 
\Level{3}{chiang2019learning}, 
\Level{1}{zhu_target-driven_2017},  
\Level{2}{hoeller_learning_2021},
\Level{1}{kadian2020sim2real}, 
\Level{4}{kahn_badgr_2021}, 
\Level{3}{shah_offline_2023}, 
\Level{3}{stachowicz_fastrlap_2023},
\Level{2}{kendall2019learning},
\Level{4}{jang2024reinforcement},
 \Level{3}{sorokin2022learning}, \Level{2}{zhang2024resilient}$^*$, \Level{3}{miki_learning_2024}, \Level{3}{sadeghi2017cad2rl}, \Level{3}{kang2019generalization}, \Level{3}{romero2023actor}
& 
\Level{3}{stein_learning_2018}, 
\Level{3}{chaplot_object_2020}, 
\Level{4}{gervet2023navigating}, 
\Level{3}{truong_rethinking_2023}, \Level{3}{truong_indoorsim--outdoorreal_2023}, \Level{2}{zhang2024resilient}$^*$, \Level{2}{hoeller2024anymal}$^*$, \Level{4}{lee2024learning}$^*$
\\ \hline
Manipulation &
\Level{2}{levine2016end},
\Level{1}{nair2020awac}, 
\Level{1}{vecerik_leveraging_2018}, 
\Level{3}{chebotar_closing_2019}, 
\Level{3}{andrychowicz2020learning}, 
\Level{3}{handa_dextreme_2023}, 
\Level{3}{nagabandi_deep_2020},
\Level{4}{qi2023general}, 
\Level{4}{chen2023visual},
\Level{2}{sievers_learning_2022},
\Level{3}{Pitz2024} &

\Level{1}{wu2023daydreamer},
\Level{3}{kalashnikov_scalable_2018}, 
\Level{2}{james_sim--real_2019}, 
\Level{2}{wang2023robot}, 
\Level{3}{kalashnikov2022scaling}, 
\Level{3}{chebotar_actionable_2021}, 
\Level{3}{lee2021beyond}, 
\Level{3}{walke_dont_2022}, 
\Level{2}{ebert2018visual},  
 \Level{3}{riedmiller_learning_2018}, 
\Level{3}{zhu_ingredients_2020},
\Level{1}{ma_vip_2022}, 
\Level{4}{chebotar_q-transformer_2023}, 
\Level{1}{nair_visual_2018},
\Level{1}{johannink_residual_2019},
\Level{4}{luo_robust_2021},
\Level{4}{zhao_offline_2022},
\Level{4}{tang2023industreal}, 
\Level{2}{abbatematteo2024composable},
\Level{3}{wu2022vat}, 
\Level{2}{matas_sim--real_2018}, 
\Level{4}{wang_one_2023},
\Level{3}{zhou2023learning},
\Level{4}{zhou2023hacman},
\Level{4}{cho2024corn} 
&
\Level{4}{mahler2019learning},
\Level{2}{zeng2018learning}, 
\Level{3}{nasiriany2022augmenting},
\Level{1}{wu_learning_2020},
\Level{3}{avigal2022speedfolding}
\\ \hline
MoMa 
& 
\Level{1}{ma2022combining}, \Level{3}{fu2023deep}, \Level{2}{fu2024humanplus}, 
\Level{3}{Hu-RSS-23}, \Level{3}{yang2023harmonic}, \Level{3}{cheng2023legs}, \Level{1}{ji2022hierarchical}, \Level{3}{ji2023dribblebot}, \Level{4}{liu2024visual}, \Level{3}{kumar2023cascaded}
& \Level{1}{wang2020learning}, \Level{3}{honerkamp2023n}, \Level{1}{sun2022fully}, \Level{2}{jauhri2022robot}, \Level{4}{wu2023m}, \Level{4}{yokoyama2023adaptive}, \Level{4}{herzog2023deep}, \Level{4}{uppal2024spin}
& \Level{4}{xiong2024adaptive} \\ \hline
HRI 
& \Level{1}{chen2017socially},
\Level{1}{everett2021collision}, 
\Level{1}{liang2021crowd},
\Level{3}{hirose2024selfi},
\Level{1}{liu2023safe}, \Level{1}{dimeas2015comanip}
& 
 \Level{3}{christen2023synh2r},
\Level{2}{christen2023learning},
\Level{3}{nair_learning_2022}, \Level{1}{reddy2018shared}, \Level{0}{schaff2020residual}
& \Level{1}{ghadirzadeh2020human}
\\ \hline
Multi-Robot Interaction &\Level{1}{chen2017decentralized}, \Level{1}{everett2018motion}, \Level{2}{fan2020distributed}, \Level{1}{han2022reinforcement}, \Level{1}{nachum2019multi}, \Level{2}{haarnoja2024learning}  &   
\Level{1}{sartoretti2019primal}
& \\ \bottomrule
\end{tabular}
\vspace{10pt}
\caption{Categorizing Literature based on Problem Formulation}
\label{tab:taxonomy-mdp}
\end{table}


\begin{table}[ht]\fontsize{8pt}{8pt}\selectfont
\centering
\begin{tabular}{R{1.7cm}P{2.6cm}P{2.1cm}P{1.8cm}P{2.8cm}}
\toprule
 & \multicolumn{2}{c}{Observation Space} & \multicolumn{2}{c}{Reward Function} \\ \cmidrule(lr){2-3} \cmidrule(lr){4-5}
Application & High-dim & Low-dim & Sparse & Dense \\ \hline
Locomotion & 
\Level{4}{miki2022learning}, \Level{4}{gangapurwala2022rloc}, \Level{4}{loquercio2023learning}, \Level{4}{agarwal2023legged}, \Level{4}{yang2023neural}, \Level{3}{cheng2023parkour}, \Level{3}{zhuang2023robot}, \Level{1}{wu2023daydreamer}, \Level{2}{duan2023learning} & 
\Level{2}{kumar2022adapting}, \Level{1}{tan2018sim}, \Level{1}{hwangbo2019learning}, \Level{2}{feng2023genloco}, 
\Level{2}{lee2019robust}, \Level{3}{yang2020multi}, \Level{4}{kumar2021rma}, \Level{4}{lee2020learning}, \Level{3}{choi2023learning}, \Level{4}{nahrendra2023dreamwaq}, \Level{1}{escontrela2022adversarial}, \Level{1}{ma2023learning}, \Level{3}{fu2022minimizing}, \Level{4}{jenelten2024dtc}, \Level{3}{yang2023cajun}, \Level{3}{smith2022legged}, \Level{3}{vollenweider2023advanced}, \Level{3}{margolis2023walk}, \Level{3}{smith2023demonstrating}, \Level{1}{siekmann2020learning}, \Level{1}{hanna2021grounded}, \Level{1}{siekmann2021sim}, \Level{3}{li2021reinforcement}, \Level{3}{siekmann2021blind}, \Level{3}{castillo2022reinforcement}, \Level{3}{radosavovic2023real}, \Level{3}{li_reinforcement_2024}, \Level{1}{hwangbo2017control}, \Level{2}{molchanov2019sim}, \Level{3}{kaufmann2022benchmark}, \Level{2}{zhang2023hover}, \Level{2}{eschmann2024learning} &
\Level{1}{hanna2021grounded} & 
\Level{2}{kumar2022adapting}, \Level{1}{tan2018sim}, \Level{1}{hwangbo2019learning}, \Level{2}{feng2023genloco}, \Level{2}{lee2019robust}, \Level{3}{yang2020multi}, \Level{4}{kumar2021rma}, \Level{4}{lee2020learning}, \Level{4}{miki2022learning}, \Level{4}{gangapurwala2022rloc}, \Level{3}{choi2023learning}, \Level{4}{nahrendra2023dreamwaq}, \Level{1}{escontrela2022adversarial}, \Level{1}{ma2023learning}, \Level{3}{fu2022minimizing}, \Level{4}{loquercio2023learning}, \Level{4}{agarwal2023legged}, \Level{4}{yang2023neural}, \Level{4}{jenelten2024dtc}, \Level{3}{yang2023cajun}, \Level{3}{smith2022legged}, \Level{3}{cheng2023parkour}, \Level{3}{zhuang2023robot}, \Level{3}{vollenweider2023advanced}, \Level{3}{margolis2023walk}, \Level{3}{smith2023demonstrating}, \Level{1}{wu2023daydreamer}, \Level{1}{siekmann2020learning}, \Level{1}{siekmann2021sim}, \Level{3}{li2021reinforcement}, \Level{3}{siekmann2021blind}, \Level{3}{castillo2022reinforcement}, \Level{2}{duan2023learning}, \Level{3}{radosavovic2023real}, \Level{3}{li_reinforcement_2024}, \Level{1}{hwangbo2017control}, \Level{2}{molchanov2019sim}, \Level{3}{kaufmann2022benchmark}, \Level{2}{zhang2023hover}, \Level{2}{eschmann2024learning}
\\ \hline
Navigation & 
\Level{1}{tai_virtual--real_2017}, 
\Level{2}{xu_benchmarking_2023}, 
\Level{3}{chiang2019learning}, 
\Level{3}{stein_learning_2018}, 
\Level{1}{zhu_target-driven_2017},  
\Level{3}{chaplot_object_2020}, 
\Level{4}{gervet2023navigating}, 
\Level{2}{hoeller_learning_2021}, 
\Level{1}{kadian2020sim2real}, 
\Level{3}{truong_rethinking_2023}, \Level{3}{truong_indoorsim--outdoorreal_2023},
\Level{4}{kahn_badgr_2021}, 
\Level{3}{shah_offline_2023}, 
\Level{3}{stachowicz_fastrlap_2023}, 
\Level{2}{kendall2019learning},
 \Level{3}{sorokin2022learning}, \Level{2}{zhang2024resilient}, \Level{2}{hoeller2024anymal}, \Level{4}{lee2024learning}, \Level{3}{miki_learning_2024}, \Level{1}{xu2024dexterous}, \Level{3}{sadeghi2017cad2rl}, \Level{3}{kang2019generalization}
& 
\Level{3}{kaufmann2023champion}, \Level{2}{rudin2022advanced}, \Level{3}{song2023reaching}, \Level{1}{williams2017information}, \Level{4}{jang2024reinforcement}, \Level{3}{he2024agile}, \Level{3}{romero2023actor}
& 
\Level{1}{zhu_target-driven_2017},  
\Level{2}{hoeller2024anymal}$^*$
& 
\Level{3}{kaufmann2023champion}, \Level{2}{rudin2022advanced}, \Level{3}{song2023reaching},
\Level{1}{tai_virtual--real_2017}, 
\Level{2}{xu_benchmarking_2023}, 
\Level{3}{chiang2019learning}, 
\Level{3}{stein_learning_2018}, 
\Level{3}{chaplot_object_2020}, 
\Level{4}{gervet2023navigating}, 
\Level{2}{hoeller_learning_2021},
\Level{1}{kadian2020sim2real}, 
 \Level{3}{truong_rethinking_2023}, \Level{3}{truong_indoorsim--outdoorreal_2023},
\Level{4}{kahn_badgr_2021}, 
\Level{3}{shah_offline_2023}, 
\Level{1}{williams2017information}, 
\Level{3}{stachowicz_fastrlap_2023}, 
\Level{2}{kendall2019learning},
\Level{4}{jang2024reinforcement},
 \Level{3}{sorokin2022learning}, \Level{2}{zhang2024resilient}, \Level{2}{hoeller2024anymal}$^*$, \Level{4}{lee2024learning}, \Level{3}{miki_learning_2024}, \Level{1}{xu2024dexterous}, \Level{3}{he2024agile}, \Level{3}{sadeghi2017cad2rl}, \Level{3}{kang2019generalization}, \Level{3}{romero2023actor}
\\ \hline
Manipulation & 
\Level{1}{wu2023daydreamer},
\Level{4}{mahler2019learning},
\Level{2}{zeng2018learning},
\Level{3}{kalashnikov_scalable_2018}, 
\Level{2}{james_sim--real_2019}, 
\Level{2}{wang2023robot}, 
\Level{2}{levine2016end},
\Level{3}{kalashnikov2022scaling}, 
\Level{3}{chebotar_actionable_2021}, 
\Level{3}{lee2021beyond}, 
\Level{3}{walke_dont_2022}, 
\Level{2}{ebert2018visual}, 
 \Level{3}{riedmiller_learning_2018}, 
\Level{3}{zhu_ingredients_2020},
\Level{1}{ma_vip_2022}, 
\Level{4}{chebotar_q-transformer_2023}, 
\Level{1}{nair_visual_2018},
\Level{4}{luo_robust_2021},
\Level{3}{wu2022vat},
\Level{2}{matas_sim--real_2018},  
\Level{1}{wu_learning_2020},
\Level{3}{avigal2022speedfolding},
\Level{4}{wang_one_2023},
\Level{4}{qi2023general},
\Level{4}{chen2023visual},
\Level{3}{zhou2023learning},
\Level{4}{zhou2023hacman},
\Level{4}{cho2024corn}
&
\Level{1}{nair2020awac}, 
\Level{3}{nasiriany2022augmenting},
\Level{1}{johannink_residual_2019},
\Level{1}{vecerik_leveraging_2018},
\Level{4}{zhao_offline_2022},
\Level{4}{tang2023industreal}, 
\Level{3}{chebotar_closing_2019}, 
\Level{2}{abbatematteo2024composable},
\Level{3}{andrychowicz2020learning}, 
\Level{3}{handa_dextreme_2023}, 
\Level{3}{nagabandi_deep_2020},
\Level{2}{sievers_learning_2022},
\Level{4}{Pitz2024}

&
\Level{1}{wu2023daydreamer},
\Level{4}{mahler2019learning},
\Level{3}{kalashnikov_scalable_2018}, 
\Level{2}{james_sim--real_2019}, 
\Level{2}{wang2023robot}, 
\Level{3}{kalashnikov2022scaling}, 
\Level{3}{chebotar_actionable_2021}, 
\Level{3}{walke_dont_2022}, 
\Level{2}{ebert2018visual}, 
\Level{1}{nair2020awac}, 
\Level{4}{chebotar_q-transformer_2023}, 
\Level{4}{luo_robust_2021},
\Level{4}{zhao_offline_2022},
\Level{2}{matas_sim--real_2018}
&
\Level{2}{zeng2018learning}, 
\Level{2}{levine2016end},
\Level{3}{lee2021beyond}, 
\Level{2}{ebert2018visual}, 
 \Level{3}{riedmiller_learning_2018}, 
\Level{3}{zhu_ingredients_2020},
\Level{1}{ma_vip_2022},  
\Level{3}{nasiriany2022augmenting},
\Level{1}{nair_visual_2018}, 
\Level{1}{johannink_residual_2019},
\Level{1}{vecerik_leveraging_2018}, 
\Level{4}{tang2023industreal}, 
\Level{3}{chebotar_closing_2019}, 
\Level{2}{abbatematteo2024composable},
\Level{3}{wu2022vat}, 
\Level{1}{wu_learning_2020},
\Level{3}{avigal2022speedfolding},
\Level{4}{wang_one_2023},
\Level{3}{andrychowicz2020learning}, 
\Level{3}{handa_dextreme_2023}, 
\Level{3}{nagabandi_deep_2020},
\Level{4}{qi2023general}, 
\Level{4}{chen2023visual},
\Level{2}{sievers_learning_2022},
\Level{4}{Pitz2024},
\Level{3}{zhou2023learning}
\Level{4}{zhou2023hacman},
\Level{4}{cho2024corn}

\\ \hline
MoMa &  
\Level{3}{Hu-RSS-23}, \Level{3}{yang2023harmonic}, \Level{3}{honerkamp2023n}, \Level{1}{sun2022fully}, \Level{4}{xiong2024adaptive}, \Level{4}{uppal2024spin}, \Level{4}{liu2024visual}, \Level{4}{wu2023m}, \Level{4}{yokoyama2023adaptive}, \Level{4}{herzog2023deep}  & 
\Level{1}{ma2022combining}, \Level{3}{fu2023deep}, \Level{1}{wang2020learning}, \Level{2}{fu2024humanplus}, \Level{3}{cheng2023legs}, \Level{1}{ji2022hierarchical}, \Level{3}{ji2023dribblebot}, \Level{2}{jauhri2022robot}, \Level{3}{kumar2023cascaded}
& \Level{4}{xiong2024adaptive}, \Level{4}{wu2023m}, \Level{4}{herzog2023deep} 
& \Level{1}{ma2022combining}, \Level{3}{fu2023deep}, \Level{1}{wang2020learning}, \Level{2}{fu2024humanplus}, \Level{3}{Hu-RSS-23}, \Level{3}{yang2023harmonic}, \Level{3}{cheng2023legs}, \Level{1}{ji2022hierarchical}, \Level{3}{ji2023dribblebot}, \Level{3}{honerkamp2023n}, \Level{1}{sun2022fully}, \Level{2}{jauhri2022robot}, \Level{4}{uppal2024spin}, \Level{4}{liu2024visual}, \Level{3}{kumar2023cascaded}, \Level{4}{yokoyama2023adaptive} \\ \hline
HRI 
& \Level{3}{christen2023synh2r}, \Level{2}{christen2023learning}, \Level{1}{liang2021crowd},
\Level{3}{hirose2024selfi},
\Level{3}{nair_learning_2022}
&
\Level{1}{ghadirzadeh2020human}, \Level{1}{chen2017socially}, \Level{1}{everett2021collision}, \Level{1}{liu2023safe}, \Level{1}{dimeas2015comanip}, \Level{1}{reddy2018shared}, \Level{0}{schaff2020residual}
& \Level{1}{ghadirzadeh2020human}
& \Level{3}{christen2023synh2r}, \Level{2}{christen2023learning},
\Level{1}{chen2017socially},
\Level{1}{everett2021collision},
\Level{1}{liang2021crowd},
\Level{3}{hirose2024selfi}, 
\Level{1}{liu2023safe},
\Level{1}{dimeas2015comanip}, 
\Level{3}{nair_learning_2022},
\Level{1}{reddy2018shared},
\Level{0}{schaff2020residual}
\\ \hline
Multi-Robot Interaction &  &\Level{1}{chen2017decentralized}, \Level{1}{everett2018motion}, \Level{2}{fan2020distributed}, \Level{1}{han2022reinforcement}, \Level{1}{sartoretti2019primal}, \Level{1}{nachum2019multi}, \Level{2}{haarnoja2024learning} & \Level{1}{chen2017decentralized}, \Level{1}{everett2018motion}&  \Level{2}{fan2020distributed}, \Level{1}{han2022reinforcement}, \Level{1}{sartoretti2019primal}, \Level{1}{nachum2019multi}, \Level{2}{haarnoja2024learning}\\ \bottomrule
\end{tabular}
\vspace{10pt}
\caption{Categorizing Literature based on Problem Formulation (Cont.)}
\label{tab:taxonomy-mdp2}
\end{table}


\begin{table}[ht]\fontsize{8pt}{8pt}\selectfont
\centering
\begin{tabular}{R{1.7cm}P{3cm}P{3cm}P{3cm}}
\toprule
 & \multicolumn{3}{c}{Simulator Usage} \\ \cmidrule(lr){2-4}
Application & Zero-shot Sim-to-Real & Few-shot Sim-to-Real & No Simulator\\ \hline
Locomotion & 
\Level{2}{kumar2022adapting}, \Level{1}{tan2018sim}, \Level{1}{hwangbo2019learning}, \Level{2}{feng2023genloco}, \Level{2}{lee2019robust}, \Level{3}{yang2020multi}, \Level{4}{kumar2021rma}, \Level{4}{lee2020learning}, \Level{4}{miki2022learning}, \Level{4}{gangapurwala2022rloc}, \Level{3}{choi2023learning}, \Level{4}{nahrendra2023dreamwaq}, \Level{1}{escontrela2022adversarial}, \Level{1}{ma2023learning}, \Level{3}{fu2022minimizing}, \Level{4}{agarwal2023legged}, \Level{4}{yang2023neural}, \Level{4}{jenelten2024dtc}, \Level{3}{yang2023cajun}, \Level{3}{cheng2023parkour}, \Level{3}{zhuang2023robot}, \Level{3}{vollenweider2023advanced}, \Level{3}{margolis2023walk}, \Level{1}{siekmann2020learning}, \Level{1}{siekmann2021sim}, \Level{3}{li2021reinforcement}, \Level{3}{siekmann2021blind}, \Level{3}{castillo2022reinforcement}, \Level{2}{duan2023learning}, \Level{3}{radosavovic2023real}, \Level{3}{li_reinforcement_2024}, \Level{1}{hwangbo2017control}, \Level{2}{molchanov2019sim}, \Level{3}{kaufmann2022benchmark}, \Level{2}{zhang2023hover}, \Level{2}{eschmann2024learning} &
\Level{4}{loquercio2023learning}, \Level{3}{smith2022legged}, \Level{1}{hanna2021grounded} &
\Level{3}{smith2023demonstrating}, \Level{1}{wu2023daydreamer}

\\ \hline
Navigation & 
\Level{2}{rudin2022advanced}, \Level{3}{song2023reaching}, 
\Level{1}{tai_virtual--real_2017}, 
\Level{2}{xu_benchmarking_2023}, 
\Level{3}{chiang2019learning}, 
\Level{3}{stein_learning_2018}, 
\Level{1}{zhu_target-driven_2017},  
\Level{3}{chaplot_object_2020}, 
\Level{4}{gervet2023navigating}, 
\Level{2}{hoeller_learning_2021}, 
\Level{1}{kadian2020sim2real}, 
\Level{3}{truong_rethinking_2023}, \Level{3}{truong_indoorsim--outdoorreal_2023},
\Level{4}{jang2024reinforcement},
 \Level{3}{sorokin2022learning}, \Level{2}{zhang2024resilient}, \Level{2}{hoeller2024anymal}, \Level{4}{lee2024learning}, \Level{3}{miki_learning_2024}, \Level{1}{xu2024dexterous}, \Level{3}{he2024agile}, \Level{3}{sadeghi2017cad2rl}, \Level{3}{romero2023actor}
& 
\Level{3}{kaufmann2023champion}, \Level{3}{kang2019generalization}
& 
\Level{4}{kahn_badgr_2021}, 
\Level{1}{williams2017information}, 
\Level{3}{stachowicz_fastrlap_2023},
\Level{2}{kendall2019learning}

\\ \hline
Manipulation & 
\Level{4}{mahler2019learning},
\Level{2}{james_sim--real_2019}, 
\Level{3}{nasiriany2022augmenting},
\Level{4}{tang2023industreal}, 
\Level{3}{wu2022vat},
\Level{2}{matas_sim--real_2018},  
\Level{1}{wu_learning_2020},
\Level{4}{wang_one_2023},
\Level{3}{andrychowicz2020learning}, 
\Level{3}{handa_dextreme_2023},
\Level{4}{qi2023general},
\Level{4}{chen2023visual},
\Level{2}{sievers_learning_2022},
\Level{4}{Pitz2024}, 
\Level{3}{zhou2023learning},
\Level{4}{zhou2023hacman},
\Level{4}{cho2024corn} 
 &
\Level{3}{lee2021beyond},
\Level{3}{chebotar_closing_2019}
 &
\Level{1}{wu2023daydreamer},
\Level{2}{zeng2018learning},
\Level{3}{kalashnikov_scalable_2018}, 
\Level{2}{wang2023robot}, 
\Level{2}{levine2016end},
\Level{3}{kalashnikov2022scaling}, 
\Level{3}{chebotar_actionable_2021}, 
\Level{3}{walke_dont_2022}, 
\Level{2}{ebert2018visual}, 
 \Level{3}{riedmiller_learning_2018}, 
\Level{3}{zhu_ingredients_2020},
\Level{1}{ma_vip_2022}, 
\Level{1}{nair2020awac}, 
\Level{4}{chebotar_q-transformer_2023}, 
\Level{1}{nair_visual_2018},
\Level{1}{johannink_residual_2019},
\Level{1}{vecerik_leveraging_2018},
\Level{4}{luo_robust_2021},
\Level{4}{zhao_offline_2022},
\Level{2}{abbatematteo2024composable},
\Level{3}{avigal2022speedfolding},
\Level{3}{nagabandi_deep_2020}
\\ \hline
MoMa & \Level{1}{ma2022combining}, \Level{3}{fu2023deep}, \Level{1}{wang2020learning}, \Level{2}{fu2024humanplus}, \Level{3}{Hu-RSS-23}, \Level{3}{yang2023harmonic}, \Level{3}{cheng2023legs}, \Level{3}{ji2023dribblebot}, \Level{3}{honerkamp2023n}, \Level{2}{jauhri2022robot},  \Level{4}{uppal2024spin}, \Level{4}{liu2024visual}, \Level{3}{kumar2023cascaded}, \Level{4}{wu2023m}, \Level{4}{yokoyama2023adaptive}
& \Level{1}{ji2022hierarchical}, \Level{4}{herzog2023deep} 
& \Level{1}{sun2022fully}, \Level{4}{xiong2024adaptive} 
\\ \hline
HRI 
& \Level{1}{ghadirzadeh2020human}, \Level{3}{christen2023synh2r}, \Level{2}{christen2023learning}, \Level{1}{chen2017socially}, \Level{1}{everett2021collision}, \Level{1}{liang2021crowd}, \Level{1}{liu2023safe}
& \Level{1}{reddy2018shared}
&  \Level{3}{hirose2024selfi}, \Level{3}{nair_learning_2022}, \Level{1}{dimeas2015comanip}
\\ \hline
Multi-Robot Interaction & \Level{1}{chen2017decentralized}, \Level{1}{everett2018motion}, \Level{2}{fan2020distributed}, \Level{1}{han2022reinforcement}, \Level{1}{sartoretti2019primal}, \Level{1}{nachum2019multi}, \Level{2}{haarnoja2024learning} & & \\ \bottomrule
\end{tabular}
\vspace{10pt}
\caption{Categorizing Literature based on Solution Approach}
\label{tab:taxonomy-solution}
\end{table}

\begin{table}[ht]\fontsize{8pt}{8pt}\selectfont
\centering
\begin{tabular}{R{1.7cm}P{2.2cm}P{2.7cm}P{2.7cm}P{1.8cm}}
\toprule
 & \multicolumn{2}{c}{Model Learning} & \multicolumn{2}{c}{Expert Usage} \\ \cmidrule(lr){2-3} \cmidrule(lr){4-5}
Application & with Model Learning & No Model \qquad Learning & No Expert & with Expert \\ \hline
Locomotion & 
\Level{1}{hwangbo2019learning}, \Level{2}{lee2019robust}, 
\Level{4}{lee2020learning}, \Level{4}{miki2022learning}, \Level{4}{gangapurwala2022rloc}, \Level{1}{ma2023learning}, \Level{3}{vollenweider2023advanced}, \Level{3}{margolis2023walk}, \Level{1}{wu2023daydreamer}, \Level{1}{hanna2021grounded} & 
\Level{2}{kumar2022adapting}, \Level{1}{tan2018sim}, \Level{2}{feng2023genloco}, \Level{3}{yang2020multi}, \Level{4}{kumar2021rma}, \Level{3}{choi2023learning}, \Level{4}{nahrendra2023dreamwaq}, \Level{1}{escontrela2022adversarial}, \Level{3}{fu2022minimizing}, \Level{4}{loquercio2023learning}, \Level{4}{agarwal2023legged}, \Level{4}{yang2023neural}, \Level{4}{jenelten2024dtc}, \Level{3}{yang2023cajun}, \Level{3}{smith2022legged}, \Level{3}{cheng2023parkour}, \Level{3}{zhuang2023robot}, \Level{3}{smith2023demonstrating}, \Level{1}{siekmann2020learning}, \Level{1}{siekmann2021sim}, \Level{3}{li2021reinforcement}, \Level{3}{siekmann2021blind}, \Level{3}{castillo2022reinforcement}, \Level{2}{duan2023learning}, \Level{3}{radosavovic2023real}, \Level{3}{li_reinforcement_2024}, \Level{1}{hwangbo2017control}, \Level{2}{molchanov2019sim}, \Level{3}{kaufmann2022benchmark}, \Level{2}{zhang2023hover}, \Level{2}{eschmann2024learning} &
\Level{1}{tan2018sim}, \Level{1}{hwangbo2019learning}, \Level{2}{lee2019robust}, \Level{3}{yang2020multi}, \Level{4}{kumar2021rma}, \Level{4}{lee2020learning},  \Level{4}{miki2022learning}, \Level{3}{choi2023learning}, \Level{4}{nahrendra2023dreamwaq}, \Level{1}{ma2023learning}, \Level{3}{fu2022minimizing}, \Level{4}{loquercio2023learning}, \Level{4}{agarwal2023legged}, \Level{4}{yang2023neural}, \Level{3}{yang2023cajun}, \Level{3}{cheng2023parkour}, \Level{3}{zhuang2023robot}, \Level{3}{margolis2023walk}, \Level{3}{smith2023demonstrating}, \Level{1}{wu2023daydreamer}, \Level{1}{hanna2021grounded}, \Level{1}{siekmann2021sim}, \Level{3}{siekmann2021blind}, \Level{3}{castillo2022reinforcement}, \Level{2}{duan2023learning}, \Level{3}{radosavovic2023real}, \Level{2}{molchanov2019sim}, \Level{3}{kaufmann2022benchmark}, \Level{2}{zhang2023hover}, \Level{2}{eschmann2024learning} &
\Level{2}{kumar2022adapting}, \Level{2}{feng2023genloco}, \Level{4}{gangapurwala2022rloc},  \Level{1}{escontrela2022adversarial}, \Level{4}{jenelten2024dtc}, \Level{3}{smith2022legged}, \Level{3}{vollenweider2023advanced}, \Level{1}{siekmann2020learning}, \Level{3}{li2021reinforcement}, \Level{3}{li_reinforcement_2024}, \Level{1}{hwangbo2017control}
\\ \hline
Navigation & 
\Level{3}{kaufmann2023champion}, \Level{2}{rudin2022advanced},
\Level{2}{xu_benchmarking_2023}*, 
\Level{4}{kahn_badgr_2021}, 
\Level{1}{williams2017information}, 
\Level{2}{zhang2024resilient}, \Level{2}{hoeller2024anymal}, \Level{4}{lee2024learning}, \Level{3}{miki_learning_2024}, \Level{3}{kang2019generalization}
& 
\Level{3}{song2023reaching},
\Level{1}{tai_virtual--real_2017}, 
\Level{2}{xu_benchmarking_2023}$^*$
\Level{3}{chiang2019learning}, 
\Level{3}{stein_learning_2018}, 
\Level{1}{zhu_target-driven_2017},  
\Level{3}{chaplot_object_2020}, 
\Level{4}{gervet2023navigating}, 
\Level{2}{hoeller_learning_2021}, 
\Level{1}{kadian2020sim2real}, 
\Level{3}{truong_rethinking_2023}, \Level{3}{truong_indoorsim--outdoorreal_2023},
\Level{3}{shah_offline_2023}, 
\Level{3}{stachowicz_fastrlap_2023}, 
\Level{2}{kendall2019learning},
\Level{4}{jang2024reinforcement},
\Level{3}{sorokin2022learning}, \Level{1}{xu2024dexterous}, \Level{3}{he2024agile}, \Level{3}{sadeghi2017cad2rl}, \Level{3}{romero2023actor}
& 
\Level{3}{kaufmann2023champion}, \Level{2}{rudin2022advanced}, \Level{3}{song2023reaching}, 
\Level{1}{tai_virtual--real_2017}, 
\Level{2}{xu_benchmarking_2023}, 
\Level{3}{chiang2019learning}, 
\Level{1}{zhu_target-driven_2017},  
\Level{3}{chaplot_object_2020}, 
\Level{4}{gervet2023navigating}, 
\Level{2}{hoeller_learning_2021},
\Level{1}{kadian2020sim2real}, 
\Level{3}{truong_rethinking_2023}, \Level{3}{truong_indoorsim--outdoorreal_2023}, 
\Level{4}{kahn_badgr_2021}, 
\Level{4}{jang2024reinforcement},
 \Level{3}{sorokin2022learning}, \Level{2}{zhang2024resilient}, \Level{2}{hoeller2024anymal}, \Level{3}{miki_learning_2024}, \Level{1}{xu2024dexterous}, \Level{3}{he2024agile}, \Level{3}{sadeghi2017cad2rl}, \Level{3}{kang2019generalization}, \Level{3}{romero2023actor}
& 
\Level{3}{stein_learning_2018}, 
\Level{3}{shah_offline_2023}, 
\Level{1}{williams2017information}, 
\Level{3}{stachowicz_fastrlap_2023}, 
\Level{2}{kendall2019learning},
\Level{4}{lee2024learning}
\\ \hline
Manipulation & 
\Level{1}{wu2023daydreamer},
\Level{2}{levine2016end},
\Level{2}{ebert2018visual},
\Level{3}{nagabandi_deep_2020},
\Level{4}{Pitz2024}&
\Level{4}{mahler2019learning},
\Level{2}{zeng2018learning},
\Level{3}{kalashnikov_scalable_2018}, 
\Level{2}{james_sim--real_2019}, 
\Level{2}{wang2023robot}, 
\Level{3}{kalashnikov2022scaling}, 
\Level{3}{chebotar_actionable_2021}, 
\Level{3}{lee2021beyond}, 
\Level{3}{walke_dont_2022}, 
 \Level{3}{riedmiller_learning_2018}, 
\Level{3}{zhu_ingredients_2020},
\Level{1}{ma_vip_2022}, 
\Level{1}{nair2020awac}, 
\Level{3}{nasiriany2022augmenting},
\Level{4}{chebotar_q-transformer_2023}, 
\Level{1}{nair_visual_2018},
\Level{1}{johannink_residual_2019},
\Level{1}{vecerik_leveraging_2018},
\Level{4}{luo_robust_2021},
\Level{4}{zhao_offline_2022},
\Level{4}{tang2023industreal}, 
\Level{3}{chebotar_closing_2019}, 
\Level{2}{abbatematteo2024composable},
\Level{3}{wu2022vat},
\Level{2}{matas_sim--real_2018},  
\Level{1}{wu_learning_2020},
\Level{3}{avigal2022speedfolding}, 
\Level{4}{wang_one_2023},
\Level{3}{andrychowicz2020learning}, 
\Level{3}{handa_dextreme_2023}, 
\Level{4}{qi2023general},
\Level{4}{chen2023visual},
\Level{2}{sievers_learning_2022},
\Level{3}{zhou2023learning},
\Level{4}{zhou2023hacman},
\Level{4}{cho2024corn} 
&
\Level{1}{wu2023daydreamer},
\Level{4}{mahler2019learning},
\Level{2}{zeng2018learning},
\Level{3}{kalashnikov_scalable_2018}, 
\Level{2}{james_sim--real_2019}, 
\Level{3}{kalashnikov2022scaling}, 
\Level{3}{chebotar_actionable_2021}, 
\Level{3}{lee2021beyond}, 
\Level{2}{ebert2018visual}, 
\Level{3}{riedmiller_learning_2018},  \Level{3}{nasiriany2022augmenting},
\Level{1}{nair_visual_2018},
\Level{4}{tang2023industreal}, 
\Level{3}{chebotar_closing_2019}, 
\Level{3}{wu2022vat},
\Level{1}{wu_learning_2020},
\Level{4}{wang_one_2023},
\Level{3}{andrychowicz2020learning}, 
\Level{3}{handa_dextreme_2023}, 
\Level{3}{nagabandi_deep_2020},
\Level{4}{qi2023general},
\Level{4}{chen2023visual},
\Level{2}{sievers_learning_2022},
\Level{4}{Pitz2024},
\Level{3}{zhou2023learning},
\Level{4}{zhou2023hacman},
\Level{4}{cho2024corn}  &

\Level{2}{wang2023robot}, 
\Level{2}{levine2016end},
\Level{3}{walke_dont_2022},  
\Level{3}{zhu_ingredients_2020},
\Level{1}{ma_vip_2022}, 
\Level{1}{nair2020awac}, 
\Level{4}{chebotar_q-transformer_2023}, 
\Level{1}{johannink_residual_2019},
\Level{1}{vecerik_leveraging_2018},
\Level{4}{luo_robust_2021},
\Level{4}{zhao_offline_2022},
\Level{2}{abbatematteo2024composable},
\Level{2}{matas_sim--real_2018},  
\Level{3}{avigal2022speedfolding}

\\ \hline
MoMa &   
& \Level{1}{ma2022combining}, \Level{3}{fu2023deep}, \Level{1}{wang2020learning}, \Level{2}{fu2024humanplus}, \Level{3}{Hu-RSS-23}, \Level{3}{yang2023harmonic}, \Level{3}{cheng2023legs}, \Level{1}{ji2022hierarchical}, \Level{3}{ji2023dribblebot}, \Level{3}{honerkamp2023n}, \Level{1}{sun2022fully}, \Level{2}{jauhri2022robot}, \Level{4}{xiong2024adaptive}, \Level{4}{uppal2024spin}, \Level{4}{liu2024visual}, \Level{3}{kumar2023cascaded}, \Level{4}{wu2023m}, \Level{4}{yokoyama2023adaptive}, \Level{4}{herzog2023deep}
&  \Level{1}{ma2022combining}, \Level{3}{fu2023deep}, \Level{1}{wang2020learning}, \Level{3}{Hu-RSS-23}, \Level{3}{yang2023harmonic}, \Level{3}{cheng2023legs}, \Level{1}{ji2022hierarchical}, \Level{3}{ji2023dribblebot}, \Level{3}{honerkamp2023n}, \Level{1}{sun2022fully}, \Level{2}{jauhri2022robot}, \Level{4}{uppal2024spin}, \Level{4}{liu2024visual}, \Level{3}{kumar2023cascaded}, \Level{4}{wu2023m}
& \Level{2}{fu2024humanplus}, \Level{4}{xiong2024adaptive}, \Level{4}{yokoyama2023adaptive}, \Level{4}{herzog2023deep} \\ \hline
HRI 
& 
\Level{3}{christen2023synh2r}, 
\Level{2}{christen2023learning},
\Level{3}{nair_learning_2022}
& \Level{1}{ghadirzadeh2020human},
\Level{1}{chen2017socially},
\Level{1}{everett2021collision},
\Level{1}{liang2021crowd},
\Level{3}{hirose2024selfi},
\Level{1}{liu2023safe},
\Level{1}{dimeas2015comanip}, 
\Level{1}{reddy2018shared},
\Level{0}{schaff2020residual}
& 
\Level{3}{christen2023synh2r}, 
\Level{2}{christen2023learning},
\Level{1}{liang2021crowd}, 
\Level{1}{liu2023safe},
\Level{1}{dimeas2015comanip}, 
\Level{1}{reddy2018shared},
\Level{0}{schaff2020residual}
& \Level{1}{ghadirzadeh2020human},
\Level{1}{chen2017socially},
\Level{1}{everett2021collision},
\Level{3}{hirose2024selfi},
\Level{3}{nair_learning_2022}

\\ \hline
Multi-Robot Interaction && \Level{1}{chen2017decentralized}, \Level{1}{everett2018motion}, \Level{2}{fan2020distributed}, \Level{1}{han2022reinforcement}, \Level{1}{sartoretti2019primal}, \Level{1}{nachum2019multi}, \Level{2}{haarnoja2024learning} & \Level{2}{fan2020distributed}, \Level{1}{han2022reinforcement}, \Level{1}{nachum2019multi} & \Level{1}{chen2017decentralized}, \Level{1}{everett2018motion}, \Level{1}{sartoretti2019primal}, \Level{2}{haarnoja2024learning} \\ \bottomrule
\end{tabular}
\vspace{10pt}
\caption{Categorizing Literature based on Solution Approach (Cont.)}
\label{tab:taxonomy-solution2}
\end{table}

\begin{table}[ht]\fontsize{8pt}{8pt}\selectfont
\centering
\begin{tabular}{R{1.7cm}P{2cm}P{2cm}P{2.6cm}P{2.8cm}}
\midrule
 & \multicolumn{4}{c}{Policy Optimization}\\ \cmidrule(lr){2-5}
Application & Planning & Offline & Off-Policy & On-Policy \\ \hline
Locomotion & 
&
&
\Level{3}{yang2020multi}, \Level{4}{gangapurwala2022rloc}$^*$, \Level{3}{smith2022legged}, \Level{3}{smith2023demonstrating}, \Level{1}{wu2023daydreamer}, \Level{2}{eschmann2024learning} &
\Level{2}{kumar2022adapting}, \Level{1}{tan2018sim}, \Level{1}{hwangbo2019learning}, \Level{2}{feng2023genloco}, \Level{2}{lee2019robust}, \Level{3}{yang2020multi}, \Level{4}{kumar2021rma}, \Level{4}{lee2020learning}, \Level{4}{miki2022learning}, \Level{4}{gangapurwala2022rloc}$^*$, \Level{3}{choi2023learning}, \Level{4}{nahrendra2023dreamwaq}, \Level{1}{escontrela2022adversarial}, \Level{1}{ma2023learning}, \Level{3}{fu2022minimizing}, \Level{4}{loquercio2023learning}, \Level{4}{agarwal2023legged}, \Level{4}{yang2023neural}, \Level{4}{jenelten2024dtc}, \Level{3}{yang2023cajun}, \Level{3}{cheng2023parkour}, \Level{3}{zhuang2023robot}, \Level{3}{vollenweider2023advanced}, \Level{3}{margolis2023walk}, \Level{1}{siekmann2020learning}, \Level{1}{hanna2021grounded}, \Level{1}{siekmann2021sim}, \Level{3}{li2021reinforcement}, \Level{3}{siekmann2021blind}, \Level{3}{castillo2022reinforcement}, \Level{2}{duan2023learning}, \Level{3}{radosavovic2023real}, \Level{3}{li_reinforcement_2024}, \Level{1}{hwangbo2017control}, \Level{2}{molchanov2019sim}, \Level{3}{kaufmann2022benchmark}, \Level{2}{zhang2023hover}
\\ \hline
Navigation & 
\Level{2}{xu_benchmarking_2023}$^*$, 
\Level{3}{stein_learning_2018}$^*$, 
\Level{4}{kahn_badgr_2021}, 
\Level{1}{williams2017information}, 
\Level{3}{kang2019generalization}, \Level{3}{romero2023actor}$^*$
& 
\Level{3}{stein_learning_2018}$^*$, 
\Level{3}{shah_offline_2023}, 
\Level{3}{stachowicz_fastrlap_2023}$^*$, 
& 
\Level{1}{tai_virtual--real_2017}, 
\Level{2}{xu_benchmarking_2023}$^*$, 
\Level{3}{chiang2019learning}, 
\Level{1}{zhu_target-driven_2017},  
\Level{3}{stachowicz_fastrlap_2023}$^*$, 
\Level{2}{kendall2019learning},
\Level{3}{sorokin2022learning}, \Level{3}{sadeghi2017cad2rl}
& 
\Level{3}{kaufmann2023champion}, \Level{2}{rudin2022advanced}, \Level{3}{song2023reaching}, 
\Level{3}{chaplot_object_2020}, 
\Level{4}{gervet2023navigating}, \Level{2}{hoeller_learning_2021}, 
\Level{1}{kadian2020sim2real}, \Level{3}{truong_rethinking_2023}, \Level{3}{truong_indoorsim--outdoorreal_2023}, 
\Level{4}{jang2024reinforcement},
\Level{2}{zhang2024resilient}, \Level{2}{hoeller2024anymal}, \Level{4}{lee2024learning}, \Level{3}{miki_learning_2024}, \Level{1}{xu2024dexterous}, \Level{3}{he2024agile}, \Level{3}{romero2023actor}$^*$
\\ \hline
Manipulation & 
\Level{2}{ebert2018visual},
\Level{3}{nagabandi_deep_2020} & 

\Level{4}{mahler2019learning},
\Level{3}{chebotar_actionable_2021}, 
\Level{1}{ma_vip_2022},
\Level{1}{nair2020awac},
\Level{4}{chebotar_q-transformer_2023},
\Level{4}{zhao_offline_2022}
&
\Level{1}{wu2023daydreamer},
\Level{2}{zeng2018learning},
\Level{3}{kalashnikov_scalable_2018}, 
\Level{2}{james_sim--real_2019}, 
\Level{2}{wang2023robot},
\Level{3}{kalashnikov2022scaling}, 
\Level{3}{lee2021beyond}, 
\Level{3}{walke_dont_2022}, 
 \Level{3}{riedmiller_learning_2018}, 
\Level{3}{zhu_ingredients_2020},
\Level{3}{nasiriany2022augmenting},
\Level{1}{nair_visual_2018},
\Level{1}{johannink_residual_2019},
\Level{1}{vecerik_leveraging_2018},
\Level{4}{luo_robust_2021},
\Level{2}{abbatematteo2024composable},
\Level{3}{wu2022vat},
\Level{2}{matas_sim--real_2018},  
\Level{1}{wu_learning_2020},
\Level{3}{avigal2022speedfolding},
\Level{4}{wang_one_2023},
\Level{2}{sievers_learning_2022},
\Level{3}{zhou2023learning},
\Level{4}{zhou2023hacman}
&
\Level{2}{levine2016end},
\Level{4}{tang2023industreal}, 
\Level{3}{chebotar_closing_2019}, 
\Level{3}{andrychowicz2020learning}, 
\Level{3}{handa_dextreme_2023}, 
\Level{4}{qi2023general},
\Level{4}{chen2023visual},
\Level{4}{Pitz2024},
\Level{4}{cho2024corn} 

\\ \hline
MoMa & 
& 
& \Level{1}{ji2022hierarchical}$^*$,  \Level{3}{honerkamp2023n}, \Level{1}{sun2022fully}, \Level{2}{jauhri2022robot}, \Level{4}{wu2023m}, \Level{4}{herzog2023deep}
& \Level{1}{ma2022combining}, \Level{3}{fu2023deep}, \Level{1}{wang2020learning}, \Level{2}{fu2024humanplus}, \Level{3}{Hu-RSS-23}, \Level{3}{yang2023harmonic}, \Level{3}{cheng2023legs}, \Level{1}{ji2022hierarchical}$^*$, \Level{3}{ji2023dribblebot}, \Level{4}{xiong2024adaptive}, \Level{4}{uppal2024spin}, \Level{4}{liu2024visual}, \Level{3}{kumar2023cascaded}, \Level{4}{yokoyama2023adaptive}
\\ \hline
HRI 
& \Level{3}{nair_learning_2022}
& \Level{3}{hirose2024selfi}$^*$
& \Level{1}{ghadirzadeh2020human},
\Level{3}{christen2023synh2r}, \Level{2}{christen2023learning}, \Level{3}{hirose2024selfi}$^*$, \Level{1}{liu2023safe}, \Level{1}{dimeas2015comanip}, \Level{1}{reddy2018shared}
& 
\Level{1}{chen2017socially},
\Level{1}{everett2021collision},
\Level{1}{liang2021crowd}, 
\Level{0}{schaff2020residual}
\\ \hline
Multi-Robot Interaction & &  &\Level{2}{haarnoja2024learning} & \Level{1}{chen2017decentralized}, \Level{1}{everett2018motion}, \Level{2}{fan2020distributed}, \Level{1}{han2022reinforcement}, \Level{1}{sartoretti2019primal},
\Level{1}{nachum2019multi} \\ \bottomrule
\end{tabular}
\vspace{10pt}
\caption{Categorizing Literature based on Solution Approach (Cont.)}
\label{tab:taxonomy-solution3}
\end{table}

\begin{table}[ht]\fontsize{8pt}{8pt}\selectfont
\centering
\begin{tabular}{R{1.7cm}P{2.4cm}P{2.6cm}P{2.2cm}P{2cm}}
\toprule
& \multicolumn{4}{c}{Policy/Model Representation}\\ \cmidrule(lr){2-5}
Application & MLP Only & CNN & RNN & Transformer \\ \hline
Locomotion & 
\Level{1}{tan2018sim}, \Level{1}{hwangbo2019learning}, \Level{2}{feng2023genloco}, \Level{2}{lee2019robust}, \Level{3}{yang2020multi}, \Level{4}{nahrendra2023dreamwaq}, \Level{1}{escontrela2022adversarial}, \Level{1}{ma2023learning}, \Level{4}{jenelten2024dtc}, \Level{3}{yang2023cajun}, \Level{3}{smith2022legged}, \Level{3}{vollenweider2023advanced}, \Level{3}{margolis2023walk}, \Level{3}{smith2023demonstrating}, \Level{1}{hanna2021grounded}, \Level{3}{li2021reinforcement}, \Level{3}{castillo2022reinforcement}, \Level{1}{hwangbo2017control}, \Level{2}{molchanov2019sim}, \Level{3}{kaufmann2022benchmark}, \Level{2}{eschmann2024learning} &
\Level{2}{kumar2022adapting}, \Level{4}{kumar2021rma}, \Level{4}{lee2020learning}, \Level{4}{gangapurwala2022rloc}, \Level{3}{fu2022minimizing}, \Level{4}{loquercio2023learning}, \Level{4}{agarwal2023legged}$^*$, \Level{4}{yang2023neural}, \Level{3}{cheng2023parkour}$^*$, \Level{1}{wu2023daydreamer}$^*$, \Level{3}{li_reinforcement_2024}, \Level{2}{zhang2023hover}
& \Level{4}{miki2022learning}, \Level{3}{choi2023learning}, \Level{4}{agarwal2023legged}$^*$, \Level{3}{cheng2023parkour}, \Level{3}{zhuang2023robot}, \Level{1}{wu2023daydreamer}$^*$, \Level{1}{siekmann2020learning}, 
 \Level{1}{siekmann2021sim}, \Level{3}{siekmann2021blind}, \Level{2}{duan2023learning} & 
\Level{3}{radosavovic2023real}

\\ \hline
Navigation & 
\Level{3}{kaufmann2023champion}, \Level{2}{rudin2022advanced}, 
\Level{3}{song2023reaching}, 
\Level{1}{tai_virtual--real_2017}, 
\Level{2}{xu_benchmarking_2023}$^*$, 
\Level{3}{chiang2019learning}, 
\Level{3}{stein_learning_2018}, 
\Level{1}{williams2017information}, 
\Level{4}{jang2024reinforcement},
\Level{1}{xu2024dexterous}, \Level{3}{he2024agile}, \Level{3}{romero2023actor}$^\mathsection$
& 
\Level{1}{zhu_target-driven_2017},  
\Level{3}{chaplot_object_2020}, 
\Level{4}{gervet2023navigating}, 
\Level{2}{hoeller_learning_2021}$^*$,
\Level{1}{kadian2020sim2real}, 
\Level{3}{truong_rethinking_2023}$^*$, \Level{3}{truong_indoorsim--outdoorreal_2023}$^*$,
\Level{3}{shah_offline_2023}, 
\Level{3}{stachowicz_fastrlap_2023}, 
\Level{2}{kendall2019learning},
  \Level{3}{sorokin2022learning}, \Level{2}{hoeller2024anymal}, \Level{4}{lee2024learning}, \Level{3}{miki_learning_2024}$^*$, \Level{3}{sadeghi2017cad2rl}, \Level{3}{kang2019generalization}$^*$
& 
\Level{2}{xu_benchmarking_2023}$^*$, 
\Level{2}{hoeller_learning_2021}$^*$,
\Level{1}{kadian2020sim2real}, 
 \Level{3}{truong_rethinking_2023}$^*$, \Level{3}{truong_indoorsim--outdoorreal_2023}$^*$, \Level{2}{zhang2024resilient}, \Level{3}{miki_learning_2024}$^*$, \Level{3}{kang2019generalization}$^*$
& 
\Level{2}{xu_benchmarking_2023}$^*$, 
\\ \hline
Manipulation &
\Level{3}{nasiriany2022augmenting},
\Level{4}{zhao_offline_2022},
\Level{3}{chebotar_closing_2019}, 
\Level{2}{abbatematteo2024composable},
\Level{3}{andrychowicz2020learning}, 
\Level{3}{handa_dextreme_2023},
\Level{3}{nagabandi_deep_2020},
\Level{2}{sievers_learning_2022},
\Level{4}{Pitz2024}$^*$,
\Level{3}{zhou2023learning}&
\Level{1}{wu2023daydreamer}$^*$,
\Level{4}{mahler2019learning},
\Level{2}{zeng2018learning},
\Level{3}{kalashnikov_scalable_2018}, 
\Level{2}{james_sim--real_2019}, 
\Level{2}{wang2023robot}, 
\Level{2}{levine2016end},
\Level{3}{kalashnikov2022scaling}, 
\Level{3}{chebotar_actionable_2021}, 
\Level{3}{walke_dont_2022},  
\Level{2}{ebert2018visual}$^*$,
\Level{3}{riedmiller_learning_2018}, 
\Level{3}{zhu_ingredients_2020},
\Level{1}{ma_vip_2022}, 
\Level{1}{nair2020awac}, 
\Level{1}{nair_visual_2018},
\Level{1}{johannink_residual_2019},
\Level{1}{vecerik_leveraging_2018},
\Level{4}{luo_robust_2021},
\Level{3}{wu2022vat},
\Level{2}{matas_sim--real_2018},
\Level{1}{wu_learning_2020},
\Level{3}{avigal2022speedfolding}, 
\Level{4}{wang_one_2023},
\Level{4}{chen2023visual},
\Level{4}{zhou2023hacman}&
\Level{1}{wu2023daydreamer}$^*$,
\Level{2}{ebert2018visual}$^*$,
\Level{4}{tang2023industreal},
\Level{4}{Pitz2024}$^*$&
\Level{3}{lee2021beyond}, 
\Level{4}{chebotar_q-transformer_2023}, 
\Level{4}{qi2023general},
\Level{4}{cho2024corn} 
\\ \hline
MoMa 
& \Level{1}{ma2022combining}, \Level{3}{fu2023deep}, \Level{1}{wang2020learning},  \Level{2}{fu2024humanplus}, \Level{3}{cheng2023legs}, \Level{1}{ji2022hierarchical}, \Level{3}{ji2023dribblebot}, \Level{2}{jauhri2022robot}, \Level{3}{kumar2023cascaded}
& \Level{3}{Hu-RSS-23}, \Level{3}{yang2023harmonic}$^*$, \Level{3}{honerkamp2023n}, \Level{1}{sun2022fully}, \Level{4}{xiong2024adaptive}, \Level{4}{uppal2024spin}$^*$, \Level{4}{liu2024visual}, \Level{4}{wu2023m}, \Level{4}{yokoyama2023adaptive}, \Level{4}{herzog2023deep}$^*$
& \Level{3}{yang2023harmonic}$^*$, \Level{4}{uppal2024spin}$^*$, \Level{4}{herzog2023deep}$^*$
\\ \hline
HRI 
& \Level{1}{chen2017socially}, 
\Level{1}{liu2023safe},
\Level{1}{dimeas2015comanip},
\Level{1}{reddy2018shared},
\Level{0}{schaff2020residual}
& \Level{3}{christen2023synh2r}, \Level{2}{christen2023learning}, \Level{1}{liang2021crowd},
\Level{3}{hirose2024selfi}, 
\Level{3}{nair_learning_2022}$^*$
& \Level{1}{ghadirzadeh2020human},
\Level{1}{everett2021collision} 
& \Level{3}{nair_learning_2022}$^*$
\\ \hline
Multi-Robot Interaction 
& \Level{1}{chen2017decentralized}, \Level{2}{fan2020distributed}, \Level{1}{nachum2019multi}, 
\Level{2}{haarnoja2024learning}
& 
& \Level{1}{everett2018motion}, \Level{1}{han2022reinforcement}, \Level{1}{sartoretti2019primal}
& \\ \bottomrule
\end{tabular}
\vspace{10pt}
\caption{Categorizing Literature based on Solution Approach (Cont.)}
\label{tab:taxonomy-solution4}
\end{table}
\end{appendices}

\end{document}